\documentclass{article}

\PassOptionsToPackage{numbers, compress}{natbib}

\usepackage[preprint]{neurips_2025}

\usepackage{tabularray} 

\usepackage{subcaption}
\usepackage{amsmath}
\usepackage[utf8]{inputenc} 
\usepackage[T1]{fontenc}    
\usepackage{hyperref}       
\usepackage{url}            
\usepackage{booktabs}       
\usepackage{amsfonts}       
\usepackage{nicefrac}       
\usepackage{microtype}      
\usepackage[dvipsnames]{xcolor}

\usepackage[utf8]{inputenc} 
\usepackage[T1]{fontenc}    
\usepackage{url}            
\usepackage{booktabs}       
\usepackage{amsfonts}       
\usepackage{nicefrac}       
\usepackage{microtype}      
\usepackage{enumitem}
\usepackage{graphicx}
\usepackage{xspace}
\usepackage{soul}
\usepackage{multirow}
\usepackage{bbold}
\usepackage{makecell}
\usepackage{hyperref}       
\usepackage{cleveref}
\usepackage{bm}

\usepackage{tikz}
\usepackage{amsmath}
\usetikzlibrary{arrows.meta, positioning}

\usepackage{pgfplots}
\usepackage{pgfplotstable} 
\usetikzlibrary{calc} 
\pgfplotsset{compat=1.18} 

\title{Linear Spatial World Models Emerge in Large Language Models}

\author{%
  Matthieu Tehenan\thanks{Equal contribution} \\
  University of Cambridge \\
  \texttt{mm2833@cam.ac.uk} \\
  \And
  Christian Bolivar Moya\footnotemark[1] \\
  Purdue University \\
  \texttt{cmoyacal@purdue.edu} \\
  \And
  Tenghai Long\footnotemark[1] \\
  Independent \\
  \texttt{theohlong@gmail.com} \\
  \And
  Guang Lin \\
  Purdue University \\
  \texttt{guanglin@purdue.edu} \\
}

\begin{document}

\maketitle

\begin{abstract}

Large language models (LLMs) have demonstrated emergent abilities across diverse tasks, raising the question of whether they acquire internal world models. In this work, we investigate whether LLMs implicitly encode linear spatial world models, which we define as linear representations of physical space and object configurations. We introduce a formal framework for spatial world models and assess whether such structure emerges in contextual embeddings. Using a synthetic dataset of object positions, we train probes to decode object positions and evaluate geometric consistency of the underlying space. We further conduct causal interventions  to test whether these spatial representations are functionally used by the model. Our results provide empirical evidence that LLMs encode linear spatial world models.     \\

\small
\begin{tblr}{colspec = {Q[c,m]|X[l,m]}, stretch = 0}
    \includegraphics[width=1.2em, keepaspectratio]{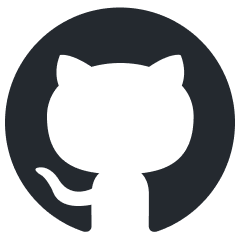}
     & \href{https://github.com/matthieu-perso/spatial_world_model}{{\textsf{linear-spatial-world-models}}} \\
\end{tblr}

\end{abstract}

\section{Introduction}

Large language models (LLMs) have demonstrated strong performance across a wide range of tasks, which suggests their abilities reflect emergent capabilities beyond their original auto-regressive training objective \citep{ wei2022emergent, bubeck2023sparks}. However, it remains an open question whether these results are simply a product of statistical imitation or whether they signal the emergence of world models. World models refer to internal representations of an environment’s dynamics, enabling an agent to simulate, predict, and regulate its behavior in response to inputs \citep{ha_world_2018, ruthenis_worldmodel_2023, yildirim_task_2023, li_emergent_2023}. There are theoretical reasons to expect the emergence of internal world models in large language models, most notably that optimal interaction with an environment would induce a system to encode a model of that environment \citep{conant1970every, francis_internal_1976}. In this sense, an internal model is not merely a beneficial by-product, but a condition for optimal behavior under certain constraints \citep{richens_robust_2024, li_emergent_2023}. 

A common critique states that LLMs capture vast numbers of surface-level correlations without acquiring any coherent model or understanding of the underlying data-generating processes, particularly due to their training on text-only data \citep{bender2020climbing, bisk2020experience}. In this view, LLMs are likened to \emph{stochastic parrots} \citep{bender_dangers_2021}, capable of producing plausible language without grasping its meaning. In contrast, an alternative perspective suggests that LLMs do, in fact, acquire internal models of the generative processes embedded in their training data, i.e. world models. These world models may emerge implicitly as a by-product of training on next-token prediction tasks, forming latent causal representations of the environment \citep{li_emergent_2023, nanda_actually_2023, ruthenis_worldmodel_2023}. 

As AI systems grow more powerful, evaluating their internal world models becomes crucial for understanding how they make decisions and for ensuring their behavior aligns with human expectations \citep{bender2021dangers, bommasani2021opportunities, ngo2023alignment, zeng2024world}. A key component of any robust world model is a spatial world model; an internal, structured representation of physical space and the objects it contains. This involves not only recognizing entities and their relationships but also representing them in a spatially coherent and compositional manner, aligned with human intuitions \citep{ha_world_2018, vafa2024evaluating}. Such spatial grounding forms the foundation of a model’s conception of its environment and is essential for building representations that are interpretable and aligned with human understanding \citep{sucholutsky_getting_2023}.

In this paper, we investigate whether large language models internally encode a spatial world model.  Specifically, we ask whether such a model is directly embedded within their contextual representations, and whether it provides coherent abstractions of space. To address this, we make the following contributions:

\begin{enumerate} \setlength{\itemsep}{0pt} 
    \item First, we formally define a spatial world model as a structured representation of discrete object locations in three-dimensional Euclidean space. 
    \item Next, we identify a linear subspace within the contextual embeddings of LLMs that corresponds to a  spatial state space, providing evidence that this subspace encodes a spatial world model. 
    \item Finally, we execute targeted causal interventions in this linear subspace to steer the model's representation of object positions, and we provide empirical evidence that this model uses this subspace for predictions.
\end{enumerate}

\section{Background}
\label{sec:background}
This section reviews prior work relevant to the emergence and analysis of internal models in large language models, focusing on three domains: interpretability, world models and spatial representations.

\paragraph{Superposition and Interpretability.} A central challenge in identifying world models within LLMs lies in the phenomenon of superposition\citep{elhage_toy_2022, olah_distributed_2023}. In such cases, multiple features are entangled within overlapping directions in the activation space, making it difficult to isolate interpretable components \citep{sharkey_taking_2022, hanni_mathematical_2024}. A first step  to disentangle these representation is to identify whether relevant information is encoded or not. Researchers have employed probing methods, which train classifiers or regressors on activations to extract information \citep{hewitt_structural_2019, belinkov2022probing}. While probes reveal what information is accessible in a model's representations, they do not confirm whether this information is functionally used \citep{ravichander_probing_2021}. To address this, causal methods from mechanistic interpretability have been developed to assess the internal use of representations \citep{olah2020zoom, rauker2023toward, bereska_mechanistic_2024}. Among these, methods like activation patching and activation steering, which replace or perturb intermediate activations, have proven effective for identifying components with causal influence on model behavior \citep{nanda_comprehensive_2022, turner_activation_2023, nanda_attribution_2023}.

\paragraph{World Models.}

The above methods have been applied to study specific world models. Empirically, there is growing evidence that LLMs can learn structured internal representations of state spaces, even in the absence of explicit supervision. Transformer models trained on structured games such as chess or Othello develop linear representations of board states and transition functions that function as emergent world models \citep{toshniwal2022chess, li2022emergent, karvonen_emergent_2024, nanda2023emergent}. These findings indicate that next-token prediction objectives can implicitly induce internal models of the environment, often organized along linear subspaces \citep{park_linear_2023, hazineh_linear_2023, nanda_actually_2023}.

\paragraph{Spatial representations.} While full spatial world models have not been explicitly uncovered, growing evidence suggests that LLMs develop structured internal representations of space and time as a by-product of next-token prediction \citep{gurnee_language_2023}. Early research showed that linguistic co-occurrence patterns encode geographic structure \citep{louwerse2009language, louwerse2012representing}, while more recent work reveals that LLMs can infer complex spatial layouts—such as maze topologies—purely from textual descriptions \citep{lietard2021language, ivanitskiy_structured_2023}. Beyond navigation, LLM representations have been found to reflect both perceptual structure in the spatial  domains, suggesting an emerging alignment with human representations \citep{patel_mapping_2022}.

\section{Spatial World Models}
\label{sec:spatial_world_models}

To assess whether a large language model implicitly encodes a spatial world model in its activations, we begin by providing a formal description.

\subsection{Definitions}

We define a spatial world model as a structure that describes the location of discrete objects in three-dimensional space. This analysis focuses solely on static spatial configurations, such as relative position, not on rotation, distances or directions.

\vspace{5pt}  
\textbf{Definition 1.} (Spatial World Model). A spatial world model \( \mathcal{W} \) is defined as a tuple:
\[
\mathcal{W} = \langle \mathbb{R}^3, \mathcal{O}, \mathcal{S} \rangle
\]
where \( \mathbb{R}^3 \) represents three-dimensional Euclidean space, \( \mathcal{O} = \{ o_1, o_2, \dots, o_n \} \) is a set of objects located in this space, and each object \( o_i \) has a position in regards to an origin \( p_i \in \mathbb{R}^3 \). The world state is given by \( \mathcal{S} = \{ p_1, p_2, \dots, p_n \} \), the set of all object positions. For example, consider a world with three objects in different locations: a bag at position \( (1.0, 2.0, 0.5) \), a car at \( (3.2, 1.0, 0.0) \), and a cat at the origin \( (0.0, 0.0, 0.0) \). Then the world state is $S$ = (1.0, 2.0, 0.5), (3.2, 1.0, 0.0),(0.0, 0.0, 0.0).

\subsection{Properties}

It follows that there exists a basis within the underlying vector space, such that all spatial relations can be reconstructed as combinations of a small set of elementary directions. These atomic spatial relations must correspond to basis directions \emph{that can be mapped to terms in natural language}. Formally, the vector space \( \mathbb{R}^3 \) in our model \( \mathcal{W} = \langle \mathbb{R}^3, \mathcal{O}, \mathcal{S} \rangle \) must satisfy two essential conditions: (1) it must admit a consistent spatial basis aligned with linguistic primitives, and (2) it must allow complex spatial relations to be expressed as linear combinations of these basis vectors.

\vspace{5pt}
\textbf{Property 1.} (Basis).  
There exists a basis of vectors \( \mathcal{B} = \{ \vec{r}_{\text{left}}, \vec{r}_{\text{above}}, \vec{r}_{\text{in front}} \} \subset \mathbb{R}^3 \), each corresponding to an atomic spatial relation in language. For instance, the words “left”, “above”, and “in front” are mapped respectively to orthogonal directions in \( \mathbb{R}^3 \), defining the primary axes of spatial reasoning. Inverse relations in language are encoded as vector negations: the vector associated with “right” is given by \( \vec{r}_{\text{right}} = -\vec{r}_{\text{left}} \), and similarly, \( \vec{r}_{\text{below}} = -\vec{r}_{\text{above}} \). This mapping preserves orthogonality among independent relations:
\[
\vec{r}_{\text{left}} \cdot \vec{r}_{\text{above}} = 0, \quad \vec{r}_{\text{left}} \cdot \vec{r}_{\text{in front}} = 0, \quad \vec{r}_{\text{above}} \cdot \vec{r}_{\text{in front}} = 0.
\]

The internal geometry of the model thus mirrors the structure of natural language descriptions of location (e.g. locative adpositions), where opposites correspond to inverses and independent relations are encoded as orthogonal directions. 

\vspace{5pt}
\textbf{Property 2.} (Composition).  
Our model $W$ represents spatial expressions as linear combinations of basis directions. For example, a spatial direction such as “above and left” corresponds to the vector sum of its atomic parts:
\[
\vec{r}_{\text{above-left}} \approx \vec{r}_{\text{above}} + \vec{r}_{\text{left}}.
\]
More generally, any composed relation is associated with a vector \( \vec{r} \in \mathbb{R}^3 \) expressible as a linear combination of the form:
\[
\vec{r} = \sum_{i} \alpha_i \vec{r}_i,
\]
where \( \vec{r}_i \in \mathcal{B} \) and \( \alpha_i \in \mathbb{R} \). 

We thus formulate the following hypothesis: if a model encodes spatial relations as linear combinations of a small set of atomic directions, then it possesses a compositional spatial world model. This structure enables the model to generate and interpret unseen spatial expressions through algebraic composition, establishing a direct mapping between spatial language (e.g. locative adpositions) and internal geometric representations.

\section{Methods}
\label{sec:methods}

To evaluate whether LLMs implicitly encode spatial world models as described above, we adopt a two-part methodology. First, we extract internal representations corresponding to world states from given LLM activations. Second, we assess whether these representations reflect the structure of the defined world model—namely: (1) the basis vectors of $\mathbb{R}^3$ and (2) coherence of object states $O$ as a composition of a these basis vectors. Our approach aims to determine whether a structured representation of space is embedded within broader contextual embeddings, and whether this representation plays a causally functional role in model behavior.

\subsection{Models and Data Selection}

\paragraph{Datasets.}  
We construct a synthetic dataset designed to represent spatial positions, composed of a set of 61 distinct objects and 6 spatial relations. Each sample in the dataset describes one or more objects placed in specific relative positions within a 2D or 3D environment. Pairs of objects are concatenated to form short natural language prompts such as: “The cup is above the table. The book is to the left of the cup.”
Each prompt is annotated with a tuple of object positions in Euclidean space, which serves as ground-truth for probing and intervention tasks. Further dataset generation details and examples are provided in Appendix \ref{appendix:dataset_construction}.
 
\paragraph{Models and Activations.}  
We perform all experiments using the \texttt{LLaMA-3.2} model family, specifically the 3B instruction-tuned variant \citep{grattafiori2024llama}. For each prompt, we extract activations from the final token prior to the period (".") using forward hooks. Unless otherwise stated, we collect the residual stream output before the final layer normalization step at layers 8, 12, and 24. In the case of multi-token object names, we average the activations across the corresponding tokens to obtain a single entity-level vector. All activations are projected to the model's hidden size, yielding an $n \times d_{\text{model}}$ matrix of activations per prompt, where $n$ is the number of datapoints. These activations are used as the input to our probing pipeline.

\subsection{Probing}

\paragraph{Probes.}  
We probe the activation vectors to detect whether spatial information is encoded, and if so, whether linearly or not. Probes are lightweight models trained to predict properties (e.g., object positions) from hidden states~\citep{hewitt_structural_2019, belinkov_probing_2021}. We train both linear and non-linear probes to assess whether spatial structure is directly encoded in the geometry of the model's representations. Formally, given activations \( \mathbf{A} \in \mathbb{R}^{n \times d_{\text{model}}} \) and targets \( \mathbf{Y} \in \mathbb{R}^{n \times d_{\text{target}}} \), we learn a linear mapping:
\[
\hat{\mathbf{W}} = \arg\min_{\mathbf{W}} \| \mathbf{Y} - \mathbf{A} \mathbf{W} \|_2^2,
\quad \text{with prediction} \quad \hat{\mathbf{Y}} = \mathbf{A} \hat{\mathbf{W}}.
\]
In parallel, we train non-linear probes \( f_\theta: \mathbb{R}^{d_model} \to \mathbb{R}^n.  \) using shallow MLP trained to minimize $ \| \mathbf{Y} - f_{\theta}(\mathbf{A}) \|_2^2.$. High linear probe performance would suggest that spatial abstraction is linearly encoded, while a significant gap between linear and non-linear probes would indicate nonlinear representations. 

\paragraph{Probes and Representation Structure.} Linear probes not only detect the presence of information, but also recover representational structure. A trained probe learns a weight vector \( \mathbf{w}_i \) for each class \( i \), with scores computed as:~\(\text{score}_i(\mathbf{h}) = \mathbf{w}_i^\top \mathbf{h}.\) Successful probing implies that class representations \( \mathbf{h} \) are linearly separable and form clusters, with hyperplanes corresponding to decision boundaries. Probe directions approximate the difference between class means:  \(\mathbf{w}_i \propto \boldsymbol{\mu}_i - \boldsymbol{\mu}_{\text{rest}},\)
where \( \boldsymbol{\mu}_i \) is the average activation for class \( i \), and \( \boldsymbol{\mu}_{\text{rest}} \) the mean of all others. Thus, each probe vector points toward the region occupied by its associated class in representation space. In this case, if relations such as ``above'' and ``below'' emerge as antipodal directions, they would reflect alignment between semantic concepts and the model’s internal geometry~\citep{makelov_sparse_2024, bricken_monosemanticity_2023}.

\subsection{Causal Experiments}
\label{sec:causal-experiments}
While probing can reveal the presence of information, it does not confirm that the representation is used by the model in a causally functional way~\citep{ravichander_probing_2021}.  We use the activation steering method ~\citep{turner_activation_2023, jorgensen_improving_2023}, where we inject direction vectors into intermediate activations and observe output shifts. Given a hidden state \( h \in \mathbb{R}^d \), we apply a steering vector \( v_r \) for relation \( r \) with scale \( \alpha \), modifying it as:
$h' = h + \alpha v_r$. We perform this intervention at selected transformer layers and evaluate whether the model’s output shifts toward the target spatial relation \( r \). Success for each relation is defined as the average correctness: \( \frac{1}{N_r} \sum_{i=1}^{N_r} \mathbb{1}[\text{steer}(x_i) \models r] \), where \( \mathbb{1}[\cdot] \) denotes lexical match. We report aggregate scores and confidence intervals.

\section{Experiments}
\label{sec:experiments}

We now turn to empirically validating the presence of a spatial world model. We study each component of the spatial world model outlined in Section \ref{sec:spatial_world_models}, starting from its most fundamental components up to its properties. We then validate these findings with causal experiments.

\subsection{Presence of a state space}

\paragraph{Motivation}
This section test the encoding of an Euclidean space $\mathbb{R}^3$, a key part of our world model tuple (Definition 1). If the model learned such a world model, each spatial relation would correspond to a specific direction or position within this space. Semantically opposed relations (e.g., “above” vs. “below”) would be encoded as approximately antipodal vectors within a shared subspace. Conversely, unrelated relations (e.g., “above” vs. “inside”) would be represented by orthogonal or nearly orthogonal vectors, indicating that there is little or no shared informational structure between them.

\paragraph{Experiments.} 

We begin by encoding our datasets using the target model and extracting the activation vectors $\mathbf{A}$ corresponding to the final token of each input. To assess whether relational labels are recoverable from the model's internal representations, we train both linear and non-linear probes on the activation vectors $\mathbf{A}$. These probes are trained to predict the original relational labels associated with each input. Next, we examine whether the relational representations form a structured subspace within the activation geometry.  In order to extract this subspace, we implement dimensionality reduction methods, such as PCA in order to disentangle the representations. Specifically, we test whether directional oppositions in the relation space, e.g., “above” versus “below”—are encoded as approximately antipodal vectors in $\mathbb{R}^3$. 


\paragraph{Results.}
Linear probes achieve near-perfect reconstruction of spatial relations, matching the performance of nonlinear probes. This indicates that the relevant relational information is linearly encoded in the model’s activation space, which provides empirical support for the linear representation hypothesis \citep{park_linear_2023}. Results are summarized in Appendix~\ref{tab:probe_results}.

\begin{table}[t]
\centering
\caption{Inverse relation alignment across layers. We report the cosine similarity and angle between each inverse pair, using both original and PCA-projected representations. Higher cosine similarity and lower angle indicate better alignment.}
\label{table:inverse_results}
\begin{tabular}{lccccc}
\toprule
\textbf{Layer} & \textbf{Relation} & \multicolumn{2}{c}{\textit{Original Space}} & \multicolumn{2}{c}{\textit{PCA-Projected Space}} \\
\cmidrule(lr){3-4} \cmidrule(lr){5-6}
& & Cosine Sim. & Angle (°)  & Cosine Sim.  & Angle (°)  \\
\midrule
\multirow{3}{*}{8} 
& above $\leftrightarrow$ below & 0.5370 & 57.52 & 0.7291 & 43.19 \\
& left $\leftrightarrow$ right & 0.8235 & 34.56 & 0.9990 & 2.54 \\
& in front $\leftrightarrow$ behind & 0.1228 & 82.95 & 0.8773 & 28.69 \\
\midrule
\multirow{3}{*}{16} 
& above $\leftrightarrow$ below & 0.3862 & 67.28 & 0.7889 & 37.92 \\
& left $\leftrightarrow$ right & 0.9656 & 15.07 & 0.9993 & 2.12 \\
& in front $\leftrightarrow$ behind & 0.0736 & 85.78 & 0.9554 & 17.18 \\
\midrule
\multirow{3}{*}{24} 
& above $\leftrightarrow$ below & 0.4465 & 63.48 & 0.9779 & 12.06 \\
& left $\leftrightarrow$ right & 0.9640 & 15.41 & 0.9969 & 4.55 \\
& in front $\leftrightarrow$ behind & 0.1130 & 83.51 & 0.9950 & 5.75 \\
\bottomrule
\end{tabular}
\end{table}

\begin{figure}[h!]
\centering
    \begin{subfigure}{0.33\textwidth}
    \centering
\pgfplotstableread[col sep=comma]{atomic_dir_layer_24_llama_8B_W_PCA_XY.csv}{\mydata}
\pgfplotstablegetrowsof{\mydata}
\pgfmathsetmacro{\maxRowIndex}{\pgfplotsretval - 1} 
\begin{tikzpicture}[
    scale=0.1,                
    square tip/.style args={#1}{ 
        -{Square[length=6.4pt, width=6.4pt, fill=#1]}
    }
]
  \draw[->, gray, line width=0.05pt] (-18.0,0) -- (18.0,0) node[right, black] {$x$};
  \draw[->, gray, line width=0.05pt] (0,-18.0) -- (0,18.0) node[above, black] {$y$};

  \foreach \t in {-15,15}{
    \draw (\t,0) -- (\t,-0.75) node[below=2pt] {\scriptsize $\t$};
    \draw (0,\t) -- (-0.75,\t) node[left=2pt]  {\scriptsize $\t$};
  }

  \pgfmathsetmacro{\numcolors}{4} 

  \foreach \idx in {0,...,\maxRowIndex} {
      \pgfplotstablegetelem{\the\numexpr\idx\relax}{x}\of\mydata \pgfmathsetmacro{\currentX}{\pgfplotsretval}
      \pgfplotstablegetelem{\the\numexpr\idx\relax}{y}\of\mydata \pgfmathsetmacro{\currentY}{\pgfplotsretval}
      \pgfplotstablegetelem{\the\numexpr\idx\relax}{label}\of\mydata \edef\currentLabel{\pgfplotsretval}

      \pgfmathsetmacro{\colorindex}{int(mod(\idx, \numcolors))}
      \ifcase\colorindex\relax
          \xdef\currentcolor{blue!70!black}%
      \or \xdef\currentcolor{red!70!black}%
      \or \xdef\currentcolor{green!60!black}%
      \or \xdef\currentcolor{orange!85!black}%
      \else \xdef\currentcolor{black}%
      \fi

      \pgfmathparse{ifthenelse(\currentX >= 0, "south", "south")} \edef\anchorpos{\pgfmathresult}
      \pgfmathparse{ifthenelse(\currentX >= 0, "-10pt", "16pt")} \edef\xshiftval{\pgfmathresult}

      \draw[color=\currentcolor, dashed, line width=2.5pt, square tip={\currentcolor}]
           (0,0) -- (\currentX,\currentY);

      \node [anchor=\anchorpos, font=\small, xshift=\xshiftval]
           at (\currentX,\currentY) {\currentLabel};
  } 
\end{tikzpicture}
    \subcaption{} 
  \end{subfigure}
  \begin{subfigure}{0.33\textwidth}
    \centering
\pgfplotstableread[col sep=comma]{atomic_dir_layer_24_llama_8B_W_PCA_XZ.csv}{\mydata}
\pgfplotstablegetrowsof{\mydata}
\pgfmathsetmacro{\maxRowIndex}{\pgfplotsretval - 1} 
\begin{tikzpicture}[
    scale=0.1,                
    square tip/.style args={#1}{ 
        -{Square[length=6.4pt, width=6.4pt, fill=#1]}
    }
]
  \draw[->, gray, line width=0.05pt] (-18.0,0) -- (18.0,0) node[right, black] {$x$};
  \draw[->, gray, line width=0.05pt] (0,-18.0) -- (0,18.0) node[above, black] {$z$};

  \foreach \t in {-15,15}{
    \draw (\t,0) -- (\t,-0.75) node[above=2pt] {\scriptsize $\t$};
    \draw (0,\t) -- (-0.75,\t) node[left=2pt]  {\scriptsize $\t$};
  }

  \pgfmathsetmacro{\numcolors}{4} 

  \foreach \idx in {0,...,\maxRowIndex} {
      \pgfplotstablegetelem{\the\numexpr\idx\relax}{x}\of\mydata \pgfmathsetmacro{\currentX}{\pgfplotsretval}
      \pgfplotstablegetelem{\the\numexpr\idx\relax}{y}\of\mydata \pgfmathsetmacro{\currentY}{\pgfplotsretval}
      \pgfplotstablegetelem{\the\numexpr\idx\relax}{label}\of\mydata \edef\currentLabel{\pgfplotsretval}

      \pgfmathsetmacro{\colorindex}{int(mod(\idx, \numcolors))}
      \ifcase\colorindex\relax
          \xdef\currentcolor{blue!70!black}%
      \or \xdef\currentcolor{red!70!black}%
      \or \xdef\currentcolor{green!60!black}%
      \or \xdef\currentcolor{orange!85!black}%
      \else \xdef\currentcolor{black}%
      \fi

      \pgfmathparse{ifthenelse(\currentY >= 0, "south", "north")} \edef\anchorpos{\pgfmathresult}
      \pgfmathparse{ifthenelse(\currentY >= 0, "5pt", "-5pt")} \edef\yshiftval{\pgfmathresult}

      \draw[color=\currentcolor, dashed, line width=2.5pt, square tip={\currentcolor}]
           (0,0) -- (\currentX,\currentY);

      \node [anchor=\anchorpos, font=\small, yshift=\yshiftval]
           at (\currentX,\currentY) {\currentLabel};
  } 
\end{tikzpicture}
    \subcaption{}
  \end{subfigure}
 \begin{subfigure}{0.325\textwidth}
    \centering
\pgfplotstableread[col sep=comma]{atomic_dir_layer_24_llama_8B_W_PCA_YZ.csv}{\mydata}
\pgfplotstablegetrowsof{\mydata}
\pgfmathsetmacro{\maxRowIndex}{\pgfplotsretval - 1} 
\begin{tikzpicture}[
    scale=0.18,                
    square tip/.style args={#1}{ 
        -{Square[length=6.4pt, width=6.4pt, fill=#1]}
    }
]
  \draw[->, gray, line width=0.01pt] (-10.0,0) -- (10.0,0) node[right, black] {$y$};
  \draw[->, gray, line width=0.01pt] (0,-10.0) -- (0,10.0) node[above, black] {$z$};

  \foreach \t in {-5,5}{
    \draw (\t,0) -- (\t,-0.5) node[below=2pt] {\scriptsize $\t$};
    \draw (0,\t) -- (-0.5,\t) node[left=2pt]  {\scriptsize $\t$};
  }

  \pgfmathsetmacro{\numcolors}{4} 

  \foreach \idx in {0,...,\maxRowIndex} {
      \pgfplotstablegetelem{\the\numexpr\idx\relax}{x}\of\mydata \pgfmathsetmacro{\currentX}{\pgfplotsretval}
      \pgfplotstablegetelem{\the\numexpr\idx\relax}{y}\of\mydata \pgfmathsetmacro{\currentY}{\pgfplotsretval}
      \pgfplotstablegetelem{\the\numexpr\idx\relax}{label}\of\mydata \edef\currentLabel{\pgfplotsretval}

      \pgfmathsetmacro{\colorindex}{int(mod(\idx, \numcolors))}
      \ifcase\colorindex\relax
          \xdef\currentcolor{blue!70!black}%
      \or \xdef\currentcolor{red!70!black}%
      \or \xdef\currentcolor{green!60!black}%
      \or \xdef\currentcolor{orange!85!black}%
      \else \xdef\currentcolor{black}%
      \fi

      \pgfmathparse{ifthenelse(\currentX >= 0, "south", "north")} \edef\anchorpos{\pgfmathresult}
      \pgfmathparse{ifthenelse(\currentX >= 0, "-15pt", "15pt")} \edef\yshiftval{\pgfmathresult}

      \draw[color=\currentcolor, dashed, line width=2.5pt, square tip={\currentcolor}]
           (0,0) -- (\currentX,\currentY);

      \node [anchor=\anchorpos, font=\small, yshift=\yshiftval]
           at (\currentX,\currentY) {\currentLabel};
  } 
\end{tikzpicture}
    \subcaption{}
  \end{subfigure}

 \caption{\small Projection on the plane of 3-D PCA vectors representing atomic spatial relations for layer 24 of \texttt{Llama-3.2-8B-Instruct} model. (a) \texttt{\{above, below, right, left\}}. (b) \texttt{\{left, right, in front, behind\}}. (c) \texttt{\{above, below, in front, behind\}}.}
    \label{fig:pca_inverse_relations}
\end{figure}
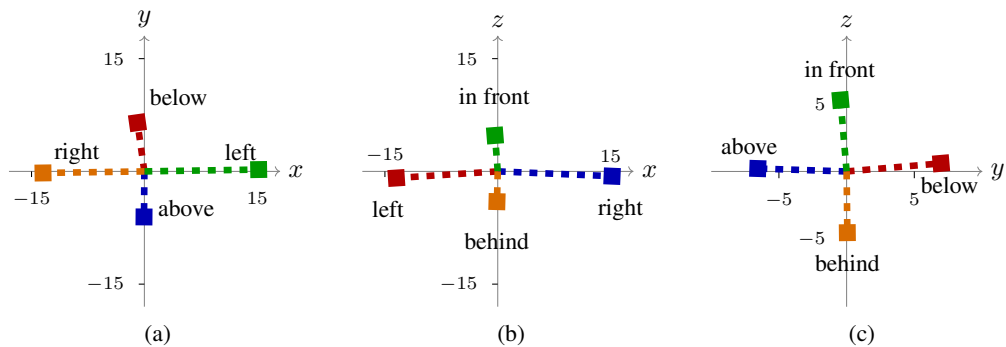


More importantly for our hypothesis, dimensionality reduction identify a lower-dimensional subspace that captures the structure of these spatial relations. Within this subspace, we observe a geometric organization: relational pairs such as \textit{above} and \textit{below} are encoded as antipodal directions, i.e., $\mathbf{w}_{\text{below}} \approx -\mathbf{w}_{\text{above}}$, while orthogonal relations such as \textit{left} and \textit{above} lie approximately at right angles. This structure is visualized in Figure~\ref{fig:pca_inverse_relations} and in 3D in Figure~\ref{fig:spatial2}. This linear organization holds across the directions of a  $\mathbb{R}^3$ basis. Notably, linear decodability is consistently observed across all tested layers (8, 16, and 24), indicating stable relational representations throughout the residual stream. Moreover, the quality of the spatial encoding improves in deeper layers, and our results suggests that this subspace becomes increasingly strucutred in downstream layers (see Table~\ref{table:inverse_results}). This suggest that the model represents spatial semantics within a subspace that is approximately isomorphic to $\mathbb{R}^3$.

\subsection{Composition}

\paragraph{Motivation.}
The above experiments indentified a subspace in which spatial relations are geometrically encoded. We now turn to evaluating its internal consistency, specifically whether the space satisfies Property 2. Our guiding question is: Do directional relations compose linearly in this space? That is, if the model encodes relations such as \textit{above} and \textit{left} as vectors, does the relation \textit{above-left} correspond to the approximate sum of those two basis vectors? If our hypothesis holds, we should observe that: (1) The vector direction of a composed relation (e.g., \textit{above and right}) approximates the sum of its constituent vector directions (e.g., \textit{above} + \textit{right}) and (2) that the angles between composed vectors and their predicted sums are minimized in a structured subspace.

\paragraph{Experiments.}
We use our dataset of labeled spatial relations, including both atomic (e.g., \textit{above}, \textit{right}) and composed relations (e.g., \textit{above} and \textit{left}, \textit{below} and \textit{right}). For each composed relation, we (1) compute the mean activation vector for each relation (e.g., $\mu_{\text{above}}$, $\mu_{\text{left}}$, $\mu_{\text{above-left}}$). We then (2) 
calculate the compositional sum: $\mu_{\text{above}} + \mu_{\text{left}}$. Finally we (3)  measure the angle between this sum and the actual mean vector of the composed relation $\mu_{\text{above-left}}$. We perform this procedure in both the original activation space and in the PCA-reduced subspace identified earlier (which we saw above captured structured spatial semantics). We run these experiments both for 2D and 3D space.

\begin{table}[t]
\centering
\small
\caption{2D Compositional Relation Metrics (Original \& PCA Spaces) — Layer 24}
\label{fig:2D_results_composition}

\begin{tabular}{l|c|c c c}
\hline
\textbf{Compositional Relation}  & \textbf{Atomic Pair} & \makecell{\textbf{Cosine}\\\textbf{Similarity}} & \makecell{\textbf{Euclidean}\\\textbf{Distance}} & \makecell{\textbf{Angle}\\(°)} \\
\hline
diag. above and right & above + right &
\makecell[l]{Orig: 0.3920\\PCA: 0.9917} &
\makecell[l]{Orig: 15.11\\PCA: 9.08} &
\makecell[l]{Orig: 66.92\\PCA: 7.39} \\
\specialrule{0.3pt}{1pt}{1pt}
diag. above and left & above + left &
\makecell[l]{Orig: 0.4083\\PCA: 0.9999} &
\makecell[l]{Orig: 14.48\\PCA: 7.52} &
\makecell[l]{Orig: 65.90\\PCA: 0.90} \\
\specialrule{0.3pt}{1pt}{1pt}
diag. below and right & below + right&
\makecell[l]{Orig: 0.3757\\PCA: 0.9917} &
\makecell[l]{Orig: 14.90\\PCA: 7.90} &
\makecell[l]{Orig: 67.93\\PCA: 7.40} \\
\specialrule{0.3pt}{1pt}{1pt}
diag. below and left & below + left &
\makecell[l]{Orig: 0.4050\\PCA: 0.9893} &
\makecell[l]{Orig: 14.85\\PCA: 8.13} &
\makecell[l]{Orig: 66.11\\PCA: 8.38} \\
\hline
\multicolumn{2}{c|}{\textbf{Mean}} &
\makecell[l]{\textbf{Orig: 0.3952}\\\textbf{PCA: 0.9931}} &
\makecell[l]{\textbf{Orig: 14.84}\\\textbf{PCA: 8.16}} &
\makecell[l]{\textbf{Orig: 66.72}\\\textbf{PCA: 6.02}} \\
\hline
\end{tabular}
\end{table}

\begin{table}[t]
\centering
\small
\caption{3D Compositional Relation Metrics (Original \& PCA Spaces) — Layer 24}
\label{fig:3D_results_composition}
\begin{tabular}{l|c|c c}
\hline
\textbf{Compositional Relation} & \textbf{Atomic Pair} & \makecell{\textbf{Cosine}\\\textbf{Similarity}} & \makecell{\textbf{Angle}\\(°)} \\
\hline
diag. above and right & above + right&
\makecell[l]{Orig: 0.3947\\PCA: 0.8799} &
\makecell[l]{Orig: 66.76\\PCA: 28.36} \\
\specialrule{0.3pt}{1pt}{1pt}
diag. above and left & above + left&
\makecell[l]{Orig: 0.3864\\PCA: 0.9183} &
\makecell[l]{Orig: 67.27\\PCA: 23.32} \\
\specialrule{0.3pt}{1pt}{1pt}
diag. below and right & below + right&
\makecell[l]{Orig: 0.3700\\PCA: 0.9187} &
\makecell[l]{Orig: 68.28\\PCA: 23.26} \\
\specialrule{0.3pt}{1pt}{1pt}
diag. below and left & below + left&
\makecell[l]{Orig: 0.3508\\PCA: 0.9139} &
\makecell[l]{Orig: 69.46\\PCA: 23.94} \\
\specialrule{0.3pt}{1pt}{1pt}
diag. above and behind & above + behind &
\makecell[l]{Orig: 0.2501\\PCA: 0.9586} &
\makecell[l]{Orig: 75.52\\PCA: 16.55} \\
\specialrule{0.3pt}{1pt}{1pt}
diag. above and in front of& above + in front of &
\makecell[l]{Orig: 0.3037\\PCA: 0.9716} &
\makecell[l]{Orig: 72.32\\PCA: 13.69} \\
\specialrule{0.3pt}{1pt}{1pt}
diag. below and behind & below + behind &
\makecell[l]{Orig: 0.3173\\PCA: 0.9977} &
\makecell[l]{Orig: 71.50\\PCA: 3.87} \\
\specialrule{0.3pt}{1pt}{1pt}
diag. below and in front of& below + in front of &
\makecell[l]{Orig: 0.2542\\PCA: 0.9785} &
\makecell[l]{Orig: 75.27\\PCA: 11.92} \\
\specialrule{0.3pt}{1pt}{1pt}
diag. left and behind & left + behind &
\makecell[l]{Orig: 0.2938\\PCA: 0.8249} &
\makecell[l]{Orig: 72.92\\PCA: 34.42} \\
\specialrule{0.3pt}{1pt}{1pt}
diag. left and in front of& left + in front of &
\makecell[l]{Orig: 0.3540\\PCA: 0.9631} &
\makecell[l]{Orig: 69.27\\PCA: 15.61} \\
\specialrule{0.3pt}{1pt}{1pt}
diag. right and behind & right + behind &
\makecell[l]{Orig: 0.4097\\PCA: 0.9824} &
\makecell[l]{Orig: 65.81\\PCA: 10.76} \\
\specialrule{0.3pt}{1pt}{1pt}
diag. right and in front of& right + in front of &
\makecell[l]{Orig: 0.3837\\PCA: 0.8453} &
\makecell[l]{Orig: 67.44\\PCA: 32.30} \\
\hline
\end{tabular}
\end{table}

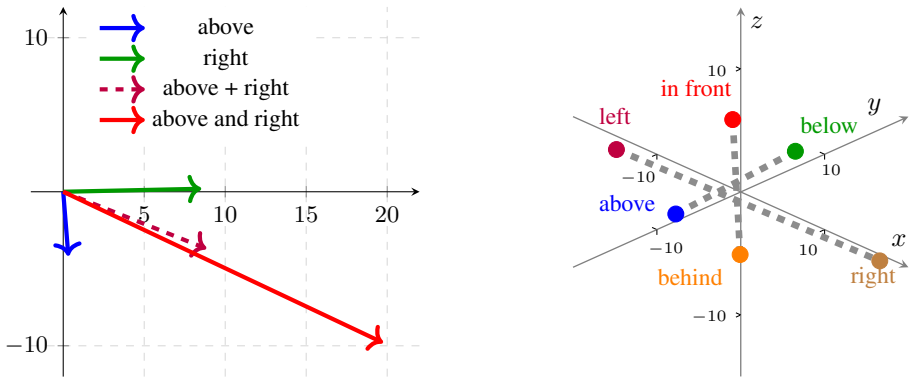
\begin{figure}[t]
    \centering
    \begin{subfigure}{0.49\textwidth}
        \centering
        \begin{tikzpicture}
  \begin{axis}[
    width=6.75cm, height=6.5cm,
    xlabel={}, ylabel={},
    xmin=-2, xmax=22,
    ymin=-12, ymax=12,
    axis lines=middle,
    grid=both,
    grid style={gray!30,dashed},
    legend style={
      at={(0.15,1.0)}, anchor=north west,
      draw=none, fill=white, font=\small
    },
    tick label style={font=\footnotesize},
  ]

    \addplot[->, ultra thick, blue] 
      coordinates { (0,0)  (0.315,-4.16) };
    \addlegendentry{above}

    \addplot[->, ultra thick, green!60!black] 
      coordinates { (0,0)  (8.521,0.1944) };
    \addlegendentry{right}

    \addplot[->, dashed, ultra thick, purple] 
      coordinates { (0,0)  (8.83,-3.69) };
    \addlegendentry{above + right}

    \addplot[->, ultra thick, red] 
      coordinates { (0,0)  (19.74,-9.80) };
    \addlegendentry{above and right}

  \end{axis}
\end{tikzpicture}
        \caption{\small 2D PCA comparison of the composition of \texttt{above} with \texttt{right} with the diagonal spatial relation: \texttt{above and right}.}
        \label{fig:spatial1}
    \end{subfigure}
    \hfill
    \begin{subfigure}{0.49\textwidth}
        \centering
        \begin{tikzpicture}
  \begin{axis}[
    view={45}{30}, 
    width=10.5cm, height=10.5cm,
    xlabel={$x$}, ylabel={$y$}, zlabel={$z$},
    xmin=-20, xmax=20,
    ymin=-20, ymax=20,
    zmin=-15, zmax=15,
    axis lines=middle,
    axis line style={line width=0.5pt, gray},
    grid=both,
    grid style={dashed,gray!30},
    tick align=outside,
    tick style={gray},
    tick style={gray},
    tick label style={font=\tiny},
    tick style={line width=0.01pt, black},
    xtick={-10,10},
    ytick={-10,10},
    ztick={-10,10},
    tick style={/pgfplots/major tick length=2pt, line width=0.4pt, draw=black},
  ]

    \addplot3[only marks, mark=*, mark size=3pt, red] 
      coordinates {(-0.4544,-0.5054,5.9011)};
    \node[red,anchor=south east] at (axis cs:0,0,7) {\small in front};

    \addplot3[only marks, mark=*, mark size=3pt, green!60!black] 
      coordinates {(-1.078,7.5954,0.6516)};
    \node[green!60!black,anchor=north west] 
      at (axis cs:0,6,5) {\small below};

    \addplot3[only marks, mark=*, mark size=3pt, blue] 
      coordinates {(-0.56,-7.185,0.2356)};
    \node[blue,anchor=north west] 
      at (axis cs:0,-18,6) {\small above};

    \addplot3[only marks, mark=*, mark size=3pt, orange] 
      coordinates {(-0.133,0.06,-5.15)};
    \node[orange,anchor=south west] 
      at (axis cs:0,-11,-5) {\small behind};

    \addplot3[only marks, mark=*, mark size=3pt, purple] 
      coordinates {(-14.6,-0.2263,-0.9526)};
    \node[purple,anchor=north] 
      at (axis cs:-15,0,3) {\small left};

    \addplot3[only marks, mark=*, mark size=3pt, brown] 
      coordinates {(16.3191,0.261,-0.6861)};
    \node[brown,anchor=south west] 
      at (axis cs:20,-8,0) {\small right};

    \draw[dashed, line width=2.5pt, gray!90] (axis cs:-0.56,-7.185,0.2356)   -- (axis cs:-1.078,7.5954,0.6516);     
    \draw[dashed, line width=2.5pt, gray!90] (axis cs:-0.4544,-0.5054,5.9011)  -- (axis cs:-0.133,0.06,-5.15); 
    \draw[dashed, line width=2.5pt, gray!90] (axis cs:-14.6,-0.2263,-0.9526)  -- (axis cs:16.3191,0.261,-0.6861);   

  \end{axis}
\end{tikzpicture}
        \caption{\small 3D PCA projection of the spatial basis vectors \texttt{above, below, in front, behind, left} and \texttt{right}, which can be composed to yield new relations.}
        \label{fig:spatial2}
    \end{subfigure}
    \caption{\small Visualization of spatial relation basis vectors and their compositional structure in PCA space.}
    \label{fig:spatial_comparison}
\end{figure}


\paragraph{Results}

Our results identify a low-dimensional subspace in which spatial relation representations exhibit clear compositional structure. Table~\ref{fig:2D_results_composition} presents the metrics comparing composed representations in 2D, obtained by vector addition of atomic relations, with directly learned compositional relations. Across all four diagonal relations, we observe high cosine similarity in the PCA space (mean = 0.9931), indicating that the composed vectors closely align with the target representations. The angular deviation in PCA space is low (mean = 6.02°), supporting the geometric consistency of these compositions. Experiments in 3D support this hypothesis, albeit with stronger angle deviations (Table~\ref{fig:3D_results_composition}). Figure~\ref{fig:spatial1} visualizes the compositional direction of the atomic vectors \textit{above} and \textit{right}, illustrating the strong alignment between composed and directly learned vectors.

\subsection{Location of Objects}

\paragraph{Motivation.} We have identified a structured subspace isomorphic to \( \mathbb{R}^3 \) within the model's activation space, corresponding to spatial relations. We now examine whether objects \( \mathcal{O} \) occupy positions \( \mathcal{S} \subseteq \mathbb{R}^3 \) within this same subspace. In other words, are objects represented at consistent locations in the spatial basis we extracted? If so, the model's internal geometry would support world states \( \mathcal{S} = \{p_1, p_2, \dots, p_n\} \), satisfying the conditions required for a spatial world model \( \mathcal{W} = \langle \mathbb{R}^3, \mathcal{O}, \mathcal{S}, T \rangle \).

\paragraph{Experiments.}

To assess whether objects occupy structured positions in the learned spatial subspace, we extract the coordinates of our objects on our basis in $R^3$. For simplicity and interpretability, we restrict our analysis to a two-dimensional plane \( \mathbb{R}^2 \subset \mathbb{R}^3 \). We project test-set activation vectors for \textit{obj1} and \textit{obj3}—taken from the residual stream before layer normalization—into this subspace, using spatial annotations (e.g., \textit{obj1 above obj2}) to select examples. This gives projected vectors \( \hat{p}_i \in \mathbb{R}^2 \) representing internal object positions. We evaluate the resulting structure via: (1) cosine similarity between the mean embedding of each object-position group and its probe direction, (2) k-means clustering purity with respect to spatial labels, and (3) variance explained by the top three PCA components to confirm compactness of the spatial subspace.

\begin{figure*}[t]
    \centering

    \begin{subfigure}[t]{0.48\textwidth}
        \centering
        \includegraphics[width=\linewidth]{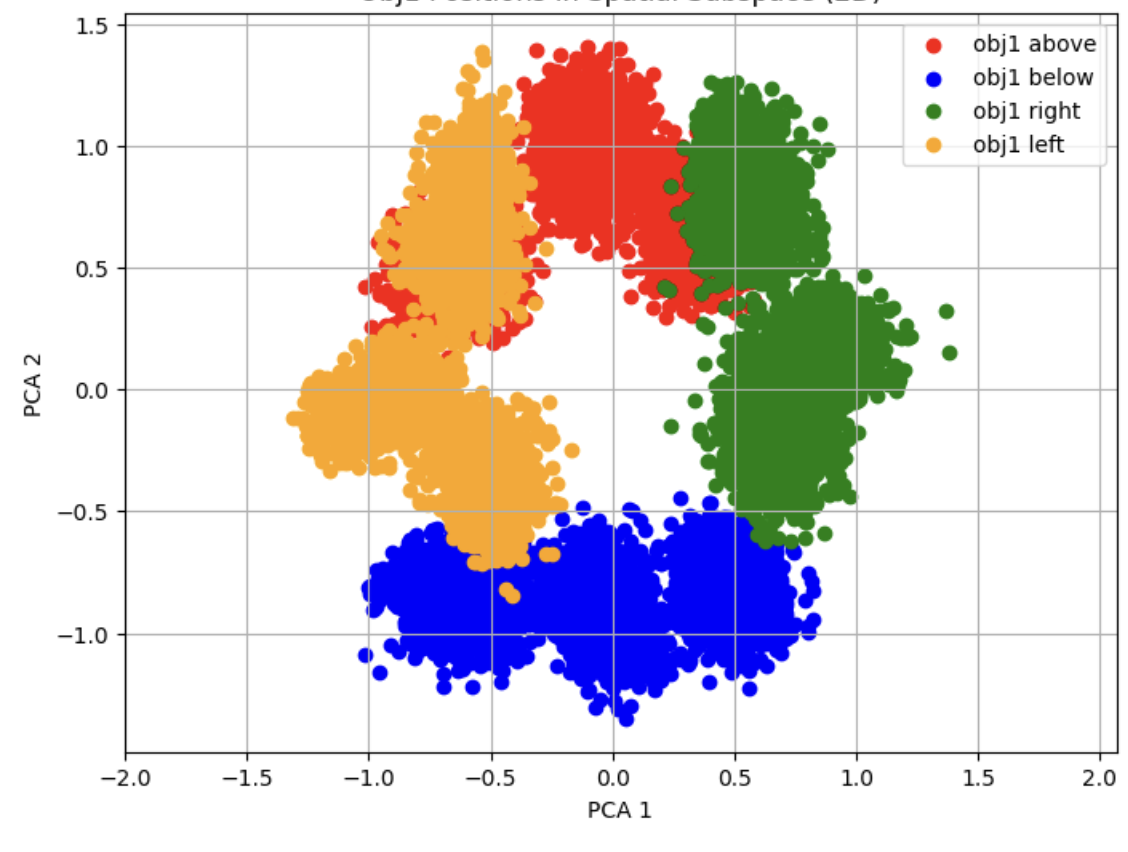}
        \caption{\small Representations of object 1. The clusters correspond to the four base relations.}
        \label{fig:spatial1}
    \end{subfigure}
    \hfill
    \begin{subfigure}[t]{0.48\textwidth}
        \centering
        \includegraphics[width=\linewidth]{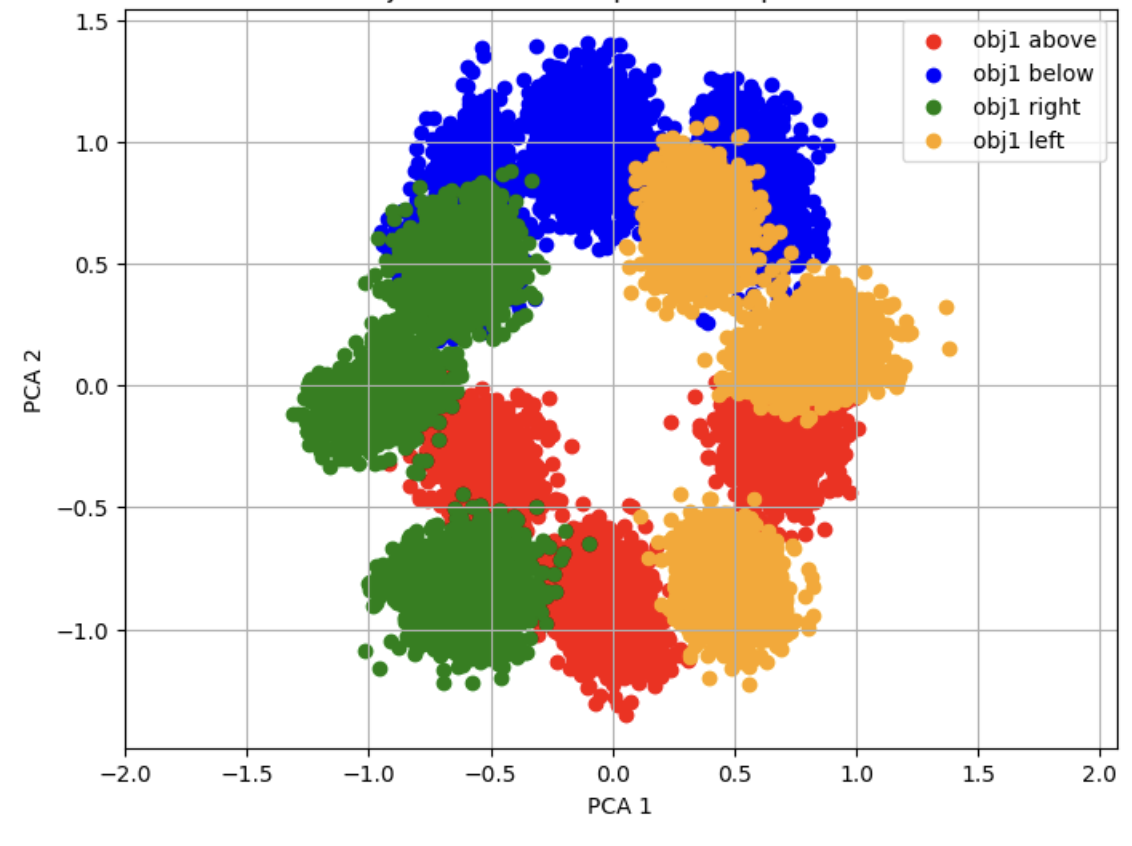}
        \caption{\small Representations of object 2 given the position of object 1.}
        \label{fig:spatial2}
    \end{subfigure}

   \caption{\small 
The relative position of two objects can be retrieved from their embeddings.
(a) Object 1 shows clear embedding clusters for each base spatial relation; 
(b) Object 2 approximates flipped configuration of object 1, yielding a mirrored embedding structure. This confirms that the learned spatial embeddings preserve relative positioning.}
    \label{fig:object_comparison}
\end{figure*}

\paragraph{Results.} We find evidence that object representations occupy consistent and distinct locations in the spatial subspace. In our last layer (layer 24), the cosine similarity between the mean embedding for \textit{obj1 above} and the projected probe direction \( \hat{w}_{\text{above}} \) reaches 0.97, indicating near-perfect alignment with the learned spatial axis. Projected object embeddings form separable clusters which achieve a purity of 77.5\%, as shown in Figure \ref{fig:object_comparison}. If obj1 is located to the left, its representation almost always partake in the left of this subspace. The spatial basis itself is highly compact, with the top three PCA directions explaining near 100\% of the variance in the probe directions. These results confirm that the model encodes object positions within the same internal frame used for spatial relations, validating the structure \( \mathcal{O} \subset \mathcal{S} \subset \mathbb{R}^2 \subset \mathcal{W} \). 

\begin{figure}[t]
    \centering
    \renewcommand{\arraystretch}{1.15} 
    \footnotesize
    \begin{tabular}{lrrrr}
        \toprule
        Relation & Succ. & Cases & Rate (\%) & 95\,\% CI (\%) \\
        \midrule
        Above    & 100 & 100 & 100.0 & [96.3, 100] \\
        Below    & 100 & 100 & 100.0 & [96.3, 100] \\
        Left     & 100 & 100 & 100.0 & [96.3, 100] \\
        Right    &  79 & 100 &  79.0 & [70.0, 85.8] \\
        In ront &  62 & 100 &  62.0 & [52.2, 70.9] \\
        Behind   &   5 & 100 &   5.0 & [2.2, 11.2] \\
        \midrule
        \textbf{Overall} & \textbf{446} & \textbf{600} & \textbf{74.3} & [70.8, 77.8] \\
        \bottomrule
    \end{tabular}
    \caption{\small Steering success by spatial relation with 95\,\% confidence intervals. We use the 3D subspace we have isolated above to steer the models.}
    \label{fig:spatial_table}
\end{figure}

\subsection{Causal Experiments}

\paragraph{Motivation.} We have identified a geometric structure \( S \subset \mathbb{R}^3 \) that appears to encode a world model, and a corresponding set of object representations \( O \) embedded within this space. Our goal is to test whether this structure is not only present but functionally used by the model for prediction (e.g., next-token prediction). We map \( S \) onto \( O \) and intervene on the positions of objects via steering operations. If the model relies on this structure, then such edits should systematically alter its behavior, e.g., modifying answers in accordance with the new object configuration.

 \paragraph{Experiments.} 
As detailed in Section~\ref{sec:causal-experiments}, we evaluate whether the spatial subspace basis of \( \mathbb{R}^3 \), identified via probing, is causally used by the model. To do so, we construct steering vectors by projecting PCA-derived spatial directions (e.g., \texttt{right}) back into the residual stream at the final token position. These vectors are injected into the model just before generation. We then ask the model to name the direction in which the object is located, and assess whether the response corresponds to our steered direction. We test across 100 samples for each of the six canonical spatial relations and their inverses (e.g., \texttt{left of}/\texttt{right of}, \texttt{above}/\texttt{below}).

\paragraph{Results.} As summarized in \ref{fig:spatial_table},  steering achieves a 74.3\,\% success rate across all relations, suggesting that we can steer via our identified subspace. Relations such as \texttt{above}, \texttt{below}, and \texttt{left of} consistently achieve near-perfect success. More challenging relations like \texttt{behind} and \texttt{in front of} yield notably lower success rates. We hypothesize that the phrase \texttt{in front of}, being distributed across multiple tokens, makes the underlying direction harder to isolate and steer. In contrast, simpler prepositions (e.g., \texttt{left of}) map more cleanly onto a single steerable direction. Overall, these results indicate that the spatial representation we have identified is used by the model in generation. 

\section{Discussion}

\paragraph{Existence of a World Model.}  
We have demonstrated the existence of a basic spatial world model embedded linearly within the representations of a large language model. The compositionality of spatial relations—captured by consistent geometric operations in a low-dimensional subspace suggests that the model internally encodes an interpretable spatial structure. This provides concrete evidence for the presence of an implicit world model, at least in the spatial domain.

\paragraph{Limitations.}
\label{sec:limitations}
While our findings point to a coherent spatial structure, this work does not investigate other essential components of a full world model, such as temporal dynamics or object permanence. In particular, movement, understood as a transition function over spatial configurations, is not addressed here. Moreover, our analysis focuses on a limited set of spatial relations. Extending the framework to cover a broader and more diverse set of spatial directions would further test the generality of the observed compositionality. Finally, although recent work suggests the shared presence of such representations across different architectures \citep{huh2024platonic}, the extent to which these findings generalize beyond the model type studied here remains an open empirical question.

\paragraph{Broader Impact.}
\label{sec:broader_impact}
 By identifying internal representations that correspond to human-interpretable spatial reasoning, our findings may help improve alignment between LLM behavior and human expectations \citep{sucholutsky_getting_2023}. This could support safer deployment of AI in applications involving grounded reasoning, such as robotics and human-computer interaction. From a societal standpoint, understanding the internal representations of  language models may help mitigate risks associated with their black-box nature, such as unintended behaviors or spurious generalizations. 
 \citep{anwar_foundational_2024, kulveit_risks_2024}



\bibliographystyle{unsrtnat}
\bibliography{arxiv}

\newpage
\appendix

\section{Experimental Details}
\label{appendix:experimental_details}

\paragraph{Dataset.} We use the full dataset of objects provided on Huggingface. All objects and spatial configurations are included without filtering or sub-sampling, ensuring complete coverage of the available examples. Examples take the form: \texttt{Object 1} is \texttt{relation} of \texttt{object 2}.

\paragraph{Probe Training.} All probes were trained using a single NVIDIA T4 GPU. We used a linear probe architecture, optimized with the Adam optimizer with a learning rate of 1 $\times$ $10^-3$ and batch size of 64. Training was performed for 100 epochs with early stopping based on validation loss. Probes were trained separately for each relation and each layer of the model. No data augmentation or regularization beyond early stopping was applied. Models converged quickly, typically within 30--35 epochs.

\section{Dataset Construction}
\label{appendix:dataset_construction}

We construct a synthetic dataset of spatial relation sentences to probe and evaluate spatial representations in language models. The dataset consists of structured natural language statements describing the spatial configuration of two objects, drawn from a controlled vocabulary.

\paragraph{Objects.} The following 61 object nouns are used to generate sentence pairs:
\texttt{"book"}, \texttt{"mug"}, \texttt{"lamp"}, \texttt{"phone"}, \texttt{"remote"}, \texttt{"cushion"}, \texttt{"plate"}, \texttt{"notebook"}, \texttt{"pen"}, \texttt{"cup"}, \texttt{"clock"}, \texttt{"chair"}, \texttt{"table"}, \texttt{"keyboard"}, \texttt{"mouse"}, \texttt{"bottle"}, \texttt{"plant"}, \texttt{"vase"}, \texttt{"wallet"}, \texttt{"bag"}, \texttt{"shoe"}, \texttt{"hat"}, \texttt{"pencil"}, \texttt{"eraser"}, \texttt{"folder"}, \texttt{"speaker"}, \texttt{"picture"}, \texttt{"mirror"}, \texttt{"pillow"}, \texttt{"blanket"}, \texttt{"carpet"}, \texttt{"painting"}, \texttt{"flower"}, \texttt{"stapler"}, \texttt{"calculator"}, \texttt{"projector"}, \texttt{"monitor"}, \texttt{"printer"}, \texttt{"scanner"}, \texttt{"microphone"}, \texttt{"camera"}, \texttt{"laptop"}, \texttt{"tablet"}, \texttt{"mousepad"}, \texttt{"desk"}, \texttt{"couch"}, \texttt{"sofa"}, \texttt{"bed"}, \texttt{"dresser"}, \texttt{"wardrobe"}, \texttt{"bookshelf"}, \texttt{"stool"}, \texttt{"bench"}, \texttt{"armchair"}, \texttt{"recliner"}, \texttt{"footstool"}, \texttt{"rug"}, \texttt{"curtain"}, \texttt{"chandelier"}, \texttt{"lamp"}, \texttt{"candle"}.

We partition these into a training set (90\%) and a disjoint test set (10\%) to ensure generalization evaluation on unseen object pairs.

\paragraph{Relations.} The spatial relations used in sentence generation are:
\texttt{"above"}, \texttt{"below"}, \texttt{"to the left of"}, \texttt{"to the right of"}, \texttt{"in front of"}, \texttt{"behind"}

\paragraph{Sentence Generation.} Sentences are constructed using the template:
\textit{``The <object\_1> is <relation> the <object\_2>.''}
This yields all pairwise combinations of object pairs and spatial relations for the training and test sets, respectively.

\section{Model Summary} \label{appendix:model_summary}
In this paper, we used \texttt{Llama-3.2-3B-Instruct} model to produce the main results in this paper \citep{grattafiori2024llama}. To validate our findings, we also conducted additional experiments with three other models: \texttt{Llama-3.2-1B-Instruct} and \texttt{Qwen3-1B}. Below, we provide a brief summary of each model.
\begin{itemize}
    \item \texttt{Llama-3.2-3B-Instruct} is a variant of Meta’s Llama 3.2 family, optimized for instruction-following. It contains approximately 3 billion parameters, composed of 28 transformer layers, and a hidden size of 3072. The model is designed to handle a range of reasoning and language tasks.
    \item \texttt{Llama-3.2-1B-Instruct} is a smaller model in the Llama 3.2 family, tailored for lightweight instruction-following tasks. It has roughly 1 billion parameters, consisting of 16 transformer layers, and a hidden size of 2048. Its compact size makes it suitable for fast inference in limited-resource environments.
    \item \texttt{Qwen3-1.7B} is an 1.7-billion-parameter language model from the Qwen3 series, developed by Alibaba \citep{qwen2, qwen2.5}. It consists of 28 transformer layers and a hidden size of 2048. The model is instruction-tuned and supports both English and Chinese, making it well-suited for multilingual and general-purpose language understanding tasks.
\end{itemize}
\section{Probing and Causal Intervention Workflows} \label{appendix:workflows}
\begin{figure}[h!]
    \centering
    \begin{tikzpicture}[
  font=\sffamily,
  every node/.style={align=center},
  box/.style={draw, minimum width=3.6cm, minimum height=1.2cm, font=\small, text width=3.4cm, inner sep=6pt},
  net/.style={draw=black, thick, fill=gray!10, minimum width=1.5cm, minimum height=1.5cm},
  arrow/.style={-{Stealth}, thick},
  node distance=1.1cm and 1.8cm
]

\node at (-5.7,2.7) {\textbf{Prompting an LLM}\\using sentences with spatial relations:\\ Object 1 is \textcolor{red}{``relation''} object 2};

\node[box, fill=cyan!20, anchor=west] (blueprompt) at (-7.8,1.2) {\textbf{Sentence:}\\e.g., the ball is \textcolor{red}{above} the table.};

\node[anchor=west] (dots) at (-6.0,0.2) {$\vdots$};

\node[box, fill=orange!30, anchor=west] (orangeprompt) at (-7.8,-1.0) {\textbf{Sentence:}\\e.g., the watch is \textcolor{red}{to the right of} the radio.};

\node at (-1.1,2.7) {\textbf{Extracting}\\ activations~$\bm{a}_\ell^{(j)}$ from\\residual stream (layer~$\ell$)};

\node[net, anchor=west] (llm1) at (-1.8,1.2) {LLM};
\node[net, anchor=west] (llm2) at (-1.8,-1.0) {LLM};

\node[anchor=west] (x1) at (0.2,1.2) {$\bm{a}^{(j)}_{\ell}$};
\node[anchor=west] (x2) at (0.2,-1.0) {$\bm{a}^{(j)}_{\ell}$};

\draw[arrow] (blueprompt.east) -- (llm1.west);
\draw[arrow] (orangeprompt.east) -- (llm2.west);

\draw[arrow] (llm1.east) -- (x1.west);
\draw[arrow] (llm2.east) -- (x2.west);

\node at (3.2,2.7) {\textbf{Training} linear probes\\to classify spatial relation~$i$,\\where $i \in$[$\text{above}$, $\text{below}, \dots$]};

\node[anchor=west] at (1.8,0.2) {$\tilde{y}^{(j)}_{\ell,i} := \bm{w}_{\ell,i}^{\top} \bm{a}^{(j)}_{\ell}$};
\end{tikzpicture}
    \caption{\small Probing Workflow.}
    \label{fig:probing}
\end{figure}
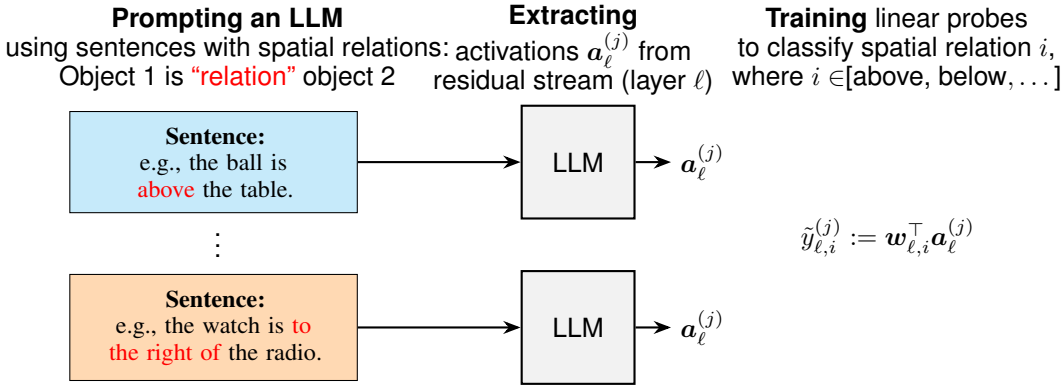

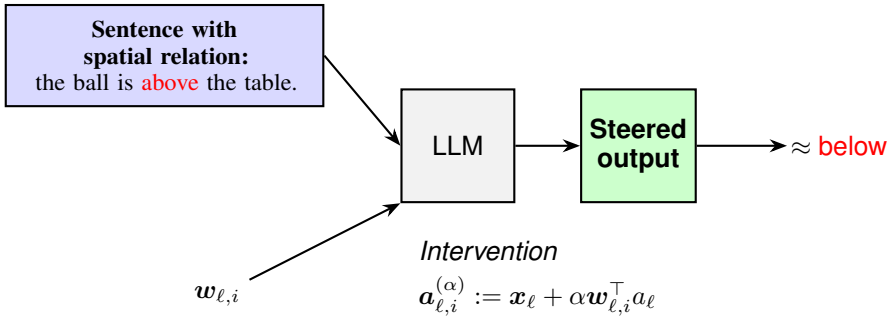
\begin{figure}[h!]
    \centering
    \begin{tikzpicture}[
  font=\sffamily,
  every node/.style={align=center},
  box/.style={draw, thick, minimum width=3.8cm, minimum height=1.3cm, text width=3.8cm, inner sep=6pt, font=\small},
  net/.style={draw=black, thick, fill=gray!10, minimum width=1.5cm, minimum height=1.5cm},
  arrow/.style={-{Stealth}, thick},
  node distance=1.2cm and 1.6cm
]

\node at (-5.5,2.8) {\textbf{Causal Intervention} using learned $i$th direction\\i.e., $\bm{w}_{\ell,i}$, which captures the spatial relation \\ between two objects in the activation space.};

\node[box, fill=blue!15] (prompt) at (-6.5,1.0) {\textbf{Sentence with spatial relation:}\\ the ball is \textcolor{red}{above} the table.};

\node (theta) at (-5.8,-2.2) {$\bm{w}_{\ell,i}$};

\node[net] (llm) at (-2.6,-0.2) {LLM};

\draw[arrow] (prompt.east) -- (llm.west);
\draw[arrow] (theta) -- (llm.south west);

\node[anchor=west] at (-3.25,-1.6) {\textit{Intervention}};
\node[anchor=west] at (-3.25,-2.2) {
$\bm{a}^{(\alpha)}_{\ell,i} := \bm{x}_{\ell} + \alpha  \bm{w}_{\ell,i} ^\top a_\ell$
};

\node[net, fill=green!20] (output) at (-0.2,-0.2) {\textbf{Steered}\\\textbf{output}};
\draw[arrow] (llm.east) -- (output.west);

\node[anchor=west] at (1.7,-0.2) {$\approx$ \textcolor{red}{below}};

\draw[arrow] (output.east) -- ++(1.2,0);

\end{tikzpicture}
    \caption{\small Causal Intervention Workflow.}
    \label{fig:intervention}
\end{figure}

Figures~\ref{fig:probing} and~\ref{fig:intervention} illustrate, respectively, the probing and causal intervention workflows used in this paper. 

\section{Additional Linear Probe Analysis for \texttt{Llama-3.2-3B-Instruction}}

\subsection{Trends in Linear Probe Accuracy Across Model Layers} \label{appendix:all_layers_acc}
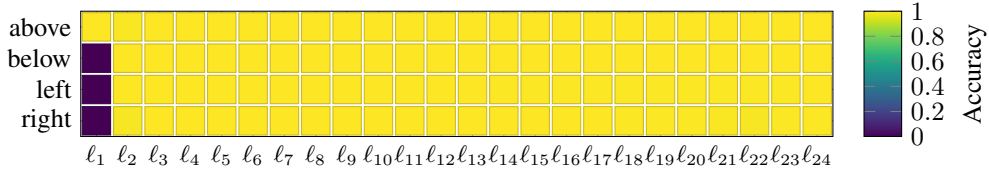
\begin{figure}[h!]
    \centering
    \begin{tikzpicture}

\begin{axis}[
    width=10cm,
    height=4cm,
    scale only axis=true,
    axis equal image,
    xticklabel style={font=\footnotesize},
    x dir=normal, y dir=reverse, 
    x=2.0cm, y=2.0cm,                
    colormap/viridis,
    colorbar,
    point meta min=0, point meta max=1, 
    colorbar style={ylabel={Accuracy}},
    xtick={1,2,3,4,5,6,7,8,9,10,11,12,13,14,15,16,17,18,19,20,21,22,23,24}, xticklabels={$\ell_1$,$\ell_2$,$\ell_3$,$\ell_4$,$\ell_5$,$\ell_6$,$\ell_7$, $\ell_8$, $\ell_9$, $\ell_{10}$,$\ell_{11}$, $\ell_{12}$, $\ell_{13}$,$\ell_{14}$, $\ell_{15}$, $\ell_{16}$,$\ell_{17}$, $\ell_{18}$, $\ell_{19}$,$\ell_{20}$, $\ell_{21}$, $\ell_{22}$,$\ell_{23}$, $\ell_{24}$},
    ytick={1,2,3,4},   yticklabels={above, below, left,right}, 
    xmin=0.5, xmax=24.5, 
    ymin=0.5, ymax=4.5, 
    enlargelimits=false,
]

\addplot[
    scatter,
    only marks,
    mark=square*,
    mark size={0.18cm + 0.25pt}, 
    point meta=explicit,
    mark options={draw=none}    
] table[
    x=x, y=y, meta=acc 
] {data.dat};

\end{axis}
\end{tikzpicture}
    \caption{\small Heatmap of classification accuracy for linear probes trained across all layers of \texttt{Llama-3.2-3B-Instruct}. The $x$-axis corresponds to model layers, from the input-adjacent bottom layer $\ell_1$ to the top layer $\ell_{24}$. The $y$-axis indicates the four atomic spatial relations: \texttt{above}, \texttt{below}, \texttt{left}, and \texttt{right}. In the heatmap, lighter shades indicate higher accuracy, while darker shades indicate lower accuracy.}
    \label{fig:all_layers_acc}
\end{figure}

In this section, Figure~\ref{fig:all_layers_acc} presents a heatmap of classification accuracy for linear probes trained and tested on 10,000 sentences of the form object 1 is \texttt{relation} of object 2 across all layers of \texttt{Llama-3.2-3B-Instruct}. The $x$-axis corresponds to model layers, from the input-adjacent bottom layer $\ell_1$ to the top layer $\ell_{24}$. The $y$-axis indicates the four atomic spatial relations: \texttt{above}, \texttt{below}, \texttt{left}, and \texttt{right}. In the heatmap, lighter shades represent higher accuracy, while darker shades indicate lower accuracy. We observe that layers above layer $\ell_3$ consistently yield high classification performance. Based on this trend, the main paper focuses its analysis on layers 8, 16, and 24.

\subsection{Additional Results for \texttt{Llama-3.2-3B-Instruction}}
\label{appendix:additionalresultssubsection}

\paragraph{Linear v.s. Nonlinear Probes.} We report probe performance across all spatial relations in Table~\ref{tab:probe_results}. Both linear and nonlinear probes achieve near-perfect accuracy at all tested layers, indicating that spatial relational information is linearly encoded in the model's activations. While high accuracies raise the risk of overfitting, the simplicity of the linear probes mitigates this concern. Evaluation on the full set of relations further ensures that the results reflect genuine structure rather than artifacts. Overall, these findings reinforce the conclusion that the model organizes spatial semantics in an accessible, low-dimensional subspace.

\begin{table}[h!]
\centering
\caption{Linear and non-linear probe performance across layers. Both probes achieve near-perfect accuracy, indicating that spatial relational information is linearly encoded throughout the model's activations.}
\label{tab:probe_results}
\begin{tabular}{lcc}
\toprule
\textbf{Layer} & \textbf{Linear Probe Accuracy} & \textbf{Non-linear Probe Accuracy} \\
\midrule
8  & 1.00 & 1.00 \\
16 & 1.00 & 1.00 \\
24 & 1.00 & 0.97 \\
\bottomrule
\end{tabular}
\end{table}

\begin{figure}[h!]
  \centering

  \begin{subfigure}[t]{0.48\textwidth}
    \centering
    \begin{tikzpicture}[scale=.145]
  \node[
    draw=gray!70,           
    fill=gray!20,           
    rectangle,
    inner sep=1pt,
    anchor=north east,
    font=\tiny,
    text=black
  ] at (20.0,14.25) {%
    \raisebox{0.1ex}{%
      \tikz{\draw[dashed,line width=3pt] (0,0) -- (0.6,0);}%
    }\,dec.\ boundary%
  };

  \draw[->] (-18.0,0) -- (18.0,0) node[right] {$x$};
  \draw[->] (0,-18.0) -- (0,18.0) node[above] {$y$};

  \foreach \t in {-15,15}{
    \draw (\t,0) -- (\t,-0.5) node[below=2pt] {\scriptsize $\t$};
    \draw (0,\t) -- (-0.5,\t) node[left=2pt]  {\scriptsize $\t$};
  }

  \pgfplotstableread[col sep=comma,header=true]{boundary_layer_24_llama_3B_above_below.csv}{\data}

  \newcommand{\getxy}[3]{%
    \pgfplotstablegetelem{#1}{x}\of\data
    \pgfmathsetmacro#2{\pgfplotsretval}
    \pgfplotstablegetelem{#1}{y}\of\data
    \pgfmathsetmacro#3{\pgfplotsretval}
  }

  \getxy{4}{\Ax}{\Ay}
  \getxy{5}{\Bx}{\By}
  \draw[line width=3pt,dashed] (\Ax,\Ay) -- (\Bx,\By);

  \def\ms{5pt}
  \foreach \i/\col/\lbl/\anc/\dx/\dy in {
     0/blue!70!black/above/west/ 2pt/ 0pt,
     1/red!70!black/below/west/ 2pt/ 0pt,
     2/gray!60/left/south/ 0pt/ 2pt,
     3/gray!60/right/south/ 0pt/ 2pt
  }{%
    \getxy{\i}{\Px}{\Py}
    \path (\Px,\Py) node[draw=\col,fill=\col,shape=rectangle,
                         minimum size=\ms,inner sep=0pt] {};
    \node[anchor=\anc,font=\small,text=\col,
          xshift=\dx,yshift=\dy]
         at (\Px,\Py) {\lbl};
  }

\end{tikzpicture}
    \subcaption{\texttt{above} $\leftrightarrow$ \texttt{below}}
  \end{subfigure}
  \hfill
  \begin{subfigure}[t]{0.48\textwidth}
    \centering
\begin{tikzpicture}[scale=.145]
  \node[
    draw=gray!70,           
    fill=gray!20,           
    rectangle,
    inner sep=1pt,
    anchor=north east,
    font=\tiny,
    text=black
  ] at (20.0,14.25) {%
    \raisebox{0.1ex}{%
      \tikz{\draw[dashed,line width=3pt] (0,0) -- (0.6,0);}%
    }\,dec.\ boundary%
  };

  \draw[->] (-18.0,0) -- (18.0,0) node[right] {$x$};
  \draw[->] (0,-18.0) -- (0,18.0) node[above] {$y$};

  \foreach \t in {-15,15}{
    \draw (\t,0) -- (\t,-0.5) node[below=2pt] {\scriptsize $\t$};
    \draw (0,\t) -- (-0.5,\t) node[left=2pt]  {\scriptsize $\t$};
  }

  \pgfplotstableread[col sep=comma,header=true]{boundary_layer_24_llama_3B_left_right.csv}{\data}

  \newcommand{\getxy}[3]{%
    \pgfplotstablegetelem{#1}{x}\of\data
    \pgfmathsetmacro#2{\pgfplotsretval}
    \pgfplotstablegetelem{#1}{y}\of\data
    \pgfmathsetmacro#3{\pgfplotsretval}
  }

  \getxy{4}{\Ax}{\Ay}
  \getxy{5}{\Bx}{\By}
  \draw[line width=3pt,dashed] (\Ax,\Ay) -- (\Bx,\By);

  \def\ms{5pt}
  \foreach \i/\col/\lbl/\anc/\dx/\dy in {
     0/gray!60/above/west/ 2pt/ 0pt,
     1/gray!60/below/west/ 2pt/ 0pt,
     2/blue!70!black/left/south/ 0pt/ 2pt,
     3/red!70!black/right/south/ 0pt/ 2pt
  }{%
    \getxy{\i}{\Px}{\Py}
    \path (\Px,\Py) node[draw=\col,fill=\col,shape=rectangle,
                         minimum size=\ms,inner sep=0pt] {};
    \node[anchor=\anc,font=\small,text=\col,
          xshift=\dx,yshift=\dy]
         at (\Px,\Py) {\lbl};
  }

\end{tikzpicture}
    \subcaption{\texttt{left} $\leftrightarrow$ \texttt{right}}
  \end{subfigure}
  \caption{\small Layer~$\ell_{24}$ decision boundaries for inverse relation pairs in \texttt{Llama-3.2-3B-Instruct}, shown in the 2D‐PCA subspace: (a) \texttt{above} $\leftrightarrow$ \texttt{below}, (b) \texttt{left} $\leftrightarrow$ \texttt{right}.}
  \label{fig:boundaries_llama_3B}
\end{figure}
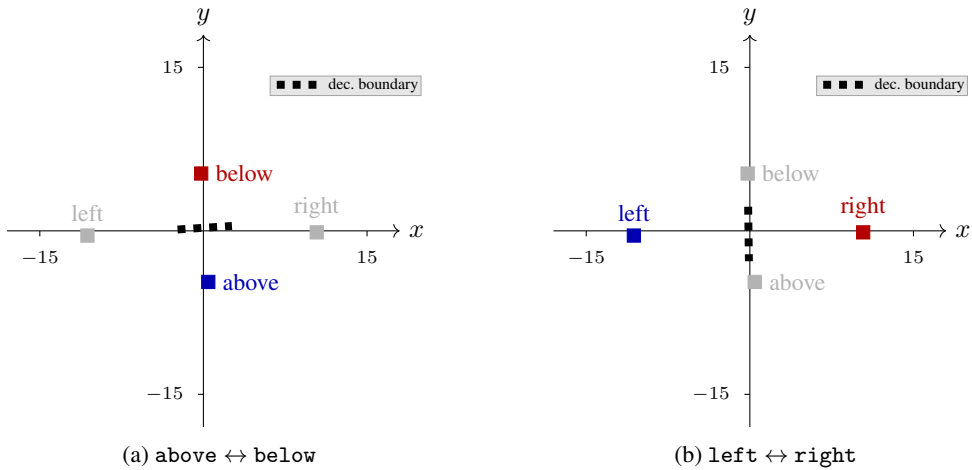

\paragraph{2D PCA Projection Additional Results.} Figure~\ref{fig:boundaries_llama_3B} illustrates decision boundaries between binary relations (\texttt{above}, \texttt{below}), and  (\texttt{left}, \texttt{right}). For completeness, we next explain the \emph{decision boundary} for feature
vectors representing inverse relations, e.g., \texttt{above} and
\texttt{below}.
Let
\( \mathbf w_{\texttt{rel1}}, \mathbf w_{\texttt{rel2}} \in \mathbb R^{d_{\text{model}}} \)
be the linear‑probe directions for two spatial relations, and let
\( \mathbf z_{\texttt{rel1}}, \mathbf z_{\texttt{rel2}} \in \mathbb R^{2} \)
denote their projections onto the PCA subspace.
Given a hidden‑state vector
\( \mathbf h \in \mathbb R^{d_{\text{model}}} \),
its PCA projection is
\( \mathbf h_{\text{proj}} \in \mathbb R^{2} \).
The binary decision boundary separating the two relations is the set of
points satisfying
\(
  (\mathbf z_{\texttt{rel1}} - \mathbf z_{\texttt{rel2}})^{\!\top}
  (\mathbf h_{\text{proj}}
        - 1/2 \bigl(\mathbf z_{\texttt{rel1}}
                        + \mathbf z_{\texttt{rel2}}\bigr)) = 0 .
\)
This boundary is a line in \( \mathbb R^{2} \) that is orthogonal to
\( \mathbf z_{\texttt{rel1}} - \mathbf z_{\texttt{rel2}} \) and passes through
the midpoint
\( \tfrac12\bigl(\mathbf z_{\texttt{rel1}} + \mathbf z_{\texttt{rel2}}\bigr) \).

\begin{figure}
    \centering
\begin{tikzpicture}
  \begin{axis}[
    width=10.0cm, height=10.0cm,
    xlabel={$x$}, ylabel={$y$},
    xmin=-1.5, xmax=1.5,
    ymin=-1.5, ymax=1.5,
    axis lines=middle,
    grid=both,
    grid style={gray!30,dashed},
  ]
    \draw[dotted] (0,0) circle (2.8cm);

    \addplot[->, line width = 2pt, yellow] 
      coordinates { (0,0)  (0.0755,-0.9971) };
      \node[yellow!90!black,anchor=north west] 
      at (axis cs:0,-1) {\small above};

    \addplot[->, line width = 2pt, blue] 
      coordinates { (0,0)  (-0.1405,0.9901) };
    \node[blue!90!black,anchor=north east] 
      at (axis cs:0,1.2) {\small below};

    \addplot[-, line width = 2pt, red, dashed] 
      coordinates { (0,0)  (-0.8946,-0.4470) };

    \addplot[->, line width = 2pt, red] 
      coordinates { (0,0)  (-0.8771,-0.4803) };
    \node[red!90!black,anchor=north east] 
      at (axis cs:-0.7,-0.5) {\small above and left};

    \addplot[-, line width = 2pt, black, dashed] 
      coordinates { (0,0)  (0.9124,-0.4094) };

    \addplot[->, line width = 2pt, black] 
      coordinates { (0,0)  (0.8957,-0.4447) };
    \node[black,anchor=north west] 
      at (axis cs:0.6,-0.5) {\small above and right};

    \addplot[-, line width = 2pt, orange, dashed] 
      coordinates { (0,0)  (-0.9046, 0.4263) };

    \addplot[->, line width = 2pt, orange] 
      coordinates { (0,0)  (-0.8854,0.4648) };
    \node[orange!90!black,anchor=north east] 
      at (axis cs:-0.6,0.7) {\small below and left};

    \addplot[-, line width = 2pt, green!60!black, dashed] 
      coordinates { (0,0)  (0.8443, 0.5359) };

    \addplot[->, line width = 2pt, green!60!black] 
      coordinates { (0,0)  (0.8846,0.4468) };
    \node[green!60!black,anchor=north west] 
      at (axis cs:0.7,0.65) {\small below and right};

    \addplot[->, line width = 2pt, purple!60!black] 
      coordinates { (0,0)  (-0.9998,-0.0223) };
    \node[purple!60!black,anchor=north east] 
      at (axis cs:-1.0,0.15) {\small left};

    \addplot[->, line width = 2pt, red!60!black] 
      coordinates { (0,0)  (0.9997, 0.0228) };
    \node[red!60!black,anchor=north west] 
      at (axis cs:0.95,0.2) {\small right};

  \end{axis}
\end{tikzpicture}
\caption{\small 2D PCA projection of atomic spatial relations: \texttt{above}, \texttt{below}, \texttt{left}, and \texttt{right}. Compositional relations are shown as dashed lines and closely align with the true diagonal directions, reflecting consistent geometric structure in the embedding space.}
    \label{fig:2D_compositional_relations}
\end{figure}
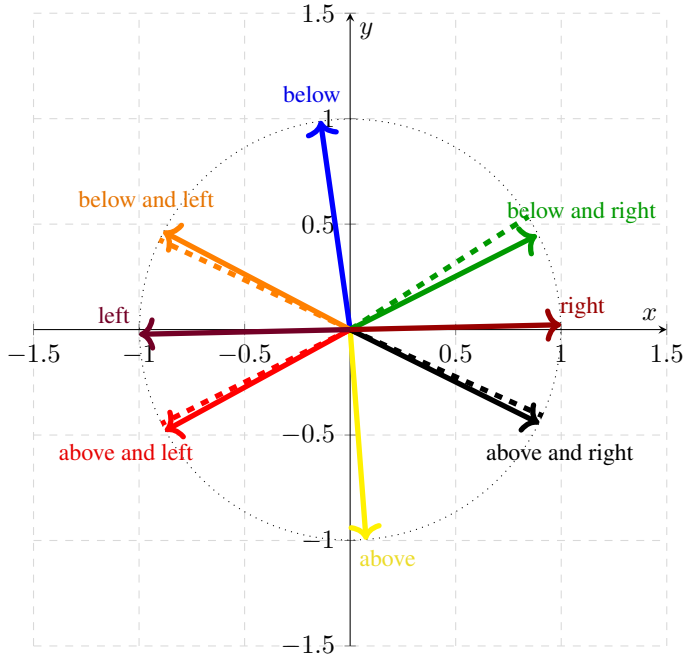

\paragraph{Additional 2D and 3D PCA Compositional Relations.}
Figure~\ref{fig:2D_compositional_relations} presents the 2D PCA projection from layer $\ell_{24}$ of the atomic spatial relations \texttt{above}, \texttt{below}, \texttt{left}, and \texttt{right}. Compositional relations are represented by dashed arrows and closely align with the true diagonal directions, indicating a consistent geometric structure in the embedding space. Table~\ref{table:3D_composition_all} reports the cosine similarities and angles computed from the 3D PCA projections at layer $\ell_{16}$ for an extended set of atomic relations—\texttt{above}, \texttt{below}, \texttt{left}, \texttt{right}, \texttt{in front of}, and \texttt{behind}—along with their corresponding compositional relations. The results show cosine similarities approaching 1; however, small angular deviations are still present.

\begin{table}[h!]
\centering
\small
\caption{3D Compositional Relation Metrics (Original \& PCA Spaces) \textemdash\ Layer 16}
\label{table:3D_composition_all}
\begin{tabular}{l|c|c c c}
\hline
\textbf{Compositional Relation} & \textbf{Atomic Pair} & \makecell{\textbf{Cosine}\\\textbf{Similarity}} & \makecell{\textbf{Euclidean}\\\textbf{Distance}} & \makecell{\textbf{Angle}\\($^{\circ}$)} \\
\hline
diag. above and right & above + right & \makecell[l]{Orig: 0.4452\\PCA: 0.8761} & \makecell[l]{Orig: 34.21\\PCA: 20.42} & \makecell[l]{Orig: 63.56\\PCA: 28.82} \\
\specialrule{0.3pt}{1pt}{1pt}
diag. above and left & above + left& \makecell[l]{Orig: 0.4286\\PCA: 0.8386} & \makecell[l]{Orig: 33.95\\PCA: 21.39} & \makecell[l]{Orig: 64.62\\PCA: 33.01} \\
\specialrule{0.3pt}{1pt}{1pt}
diag. below and right & below + right& \makecell[l]{Orig: 0.3876\\PCA: 0.8239} & \makecell[l]{Orig: 34.97\\PCA: 21.75} & \makecell[l]{Orig: 67.19\\PCA: 34.53} \\
\specialrule{0.3pt}{1pt}{1pt}
diag. below and left & below + left& \makecell[l]{Orig: 0.4270\\PCA: 0.8334} & \makecell[l]{Orig: 35.04\\PCA: 22.29} & \makecell[l]{Orig: 64.73\\PCA: 33.55} \\
\specialrule{0.3pt}{1pt}{1pt}
diag. above and behind & above + behind & \makecell[l]{Orig: 0.4092\\PCA: 0.9845} & \makecell[l]{Orig: 27.20\\PCA: 8.81} & \makecell[l]{Orig: 65.85\\PCA: 10.12} \\
\specialrule{0.3pt}{1pt}{1pt}
diag. above and in front of& above + in front of & \makecell[l]{Orig: 0.4503\\PCA: 0.9813} & \makecell[l]{Orig: 28.02\\PCA: 12.72} & \makecell[l]{Orig: 63.24\\PCA: 11.09} \\
\specialrule{0.3pt}{1pt}{1pt}
diag. below and behind & below + behind & \makecell[l]{Orig: 0.4470\\PCA: 0.9971} & \makecell[l]{Orig: 25.36\\PCA: 7.21} & \makecell[l]{Orig: 63.45\\PCA: 4.35} \\
\specialrule{0.3pt}{1pt}{1pt}
diag. below and in front of& below + in front of & \makecell[l]{Orig: 0.3892\\PCA: 0.9907} & \makecell[l]{Orig: 28.00\\PCA: 10.75} & \makecell[l]{Orig: 67.10\\PCA: 7.83} \\
\specialrule{0.3pt}{1pt}{1pt}
diag. left and behind & left + behind & \makecell[l]{Orig: 0.4373\\PCA: 0.9248} & \makecell[l]{Orig: 32.04\\PCA: 16.75} & \makecell[l]{Orig: 64.07\\PCA: 22.36} \\
\specialrule{0.3pt}{1pt}{1pt}
diag. left and in front of& left + in front of & \makecell[l]{Orig: 0.4311\\PCA: 0.9460} & \makecell[l]{Orig: 32.49\\PCA: 14.97} & \makecell[l]{Orig: 64.46\\PCA: 18.91} \\
\specialrule{0.3pt}{1pt}{1pt}
diag. right and behind & right + behind & \makecell[l]{Orig: 0.4695\\PCA: 0.9532} & \makecell[l]{Orig: 32.09\\PCA: 15.34} & \makecell[l]{Orig: 62.00\\PCA: 17.59} \\
\specialrule{0.3pt}{1pt}{1pt}
diag. right and in front of& right + in front of & \makecell[l]{Orig: 0.4097\\PCA: 0.9306} & \makecell[l]{Orig: 34.16\\PCA: 17.50} & \makecell[l]{Orig: 65.81\\PCA: 21.48} \\
\hline
\end{tabular}
\end{table}

\section{Extending Linear Probe Analysis to Other Models} \label{appendix:other_models_probes}
To validate our findings, we perform additional experiments using three other models: \texttt{Llama-3.2-1B-Instruct} and \texttt{Qwen3-1.7B}. For each model, we train and test linear probes \( W \in \mathbb{R}^{4 \times d_{\text{model}}} \) using 100,000 sentences of the form “object 1 is \texttt{relation} of object 2,” where the relation is one of the four atomic spatial relations: \texttt{above}, \texttt{below}, \texttt{left}, or \texttt{right}.

To analyze the geometric structure of these relations, we apply Principal Component Analysis (PCA) to the corresponding linear probe directions. Let \( \{\mathbf{w}_{\texttt{above}}, \mathbf{w}_{\texttt{below}}, \mathbf{w}_{\texttt{left}}, \mathbf{w}_{\texttt{right}} \} \subset \mathbb{R}^{d_{\text{model}}} \) denote the normalized probe vectors for each relation, where \( d_{\text{model}} \) is the dimensionality of the model’s hidden representation space. We stack these into a matrix \( D \in \mathbb{R}^{4 \times d_{\text{model}}} \), center the rows, and compute the empirical covariance matrix \( \Sigma = \frac{1}{4} D^\top D \). We then extract the top two eigenvectors \( \mathbf{v}_1, \mathbf{v}_2 \in \mathbb{R}^{d_{\text{model}}} \), which span the principal two-dimensional subspace. The PCA projection of each probe vector \( \mathbf{w}_i \) is given by:
\[
\mathbf{z}_i = \begin{bmatrix}
\langle \mathbf{w}_i, \mathbf{v}_1 \rangle \\
\langle \mathbf{w}_i, \mathbf{v}_2 \rangle
\end{bmatrix} \in \mathbb{R}^2.
\]
This corresponds to the orthogonal projection of the probe directions onto the plane of maximal variance. The resulting 2D configuration illustrates how each model geometrically organizes these core spatial relations.

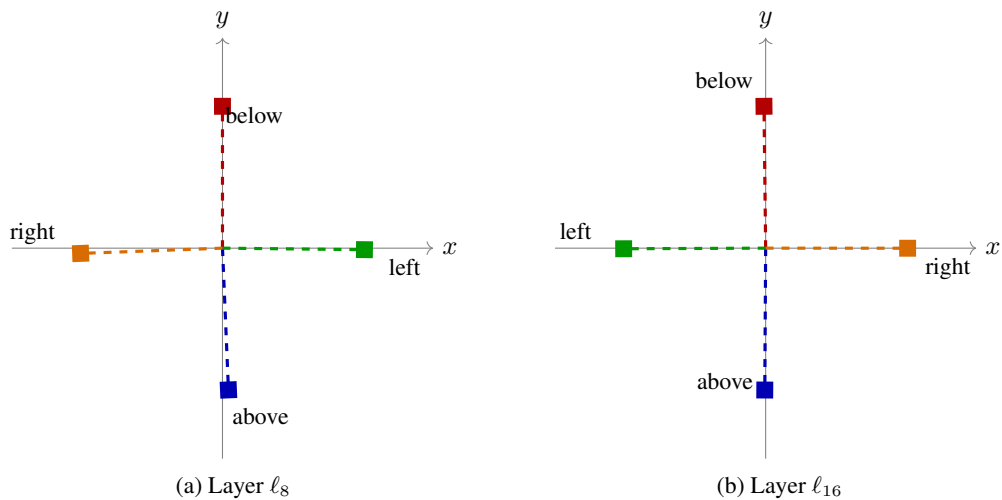
\begin{figure}[h!]
\centering
    \begin{subfigure}{0.48\textwidth}
    \centering
\pgfplotstableread[col sep=comma]{atomic_dir_layer_8_llama_1B_W_PCA.csv}{\mydata}
\pgfplotstablegetrowsof{\mydata}
\pgfmathsetmacro{\maxRowIndex}{\pgfplotsretval - 1} 
\begin{tikzpicture}[
    scale=2,                
    square tip/.style args={#1}{ 
        -{Square[length=6.4pt, width=6.4pt, fill=#1]}
    }
]
  \draw[->, gray] (-1.4,0) -- (1.4,0) node[right, black] {$x$};
  \draw[->, gray] (0,-1.4) -- (0,1.4) node[above, black] {$y$};

  \pgfmathsetmacro{\numcolors}{4} 

  \foreach \idx in {0,...,\maxRowIndex} {
      \pgfplotstablegetelem{\the\numexpr\idx\relax}{x}\of\mydata \pgfmathsetmacro{\currentX}{\pgfplotsretval}
      \pgfplotstablegetelem{\the\numexpr\idx\relax}{y}\of\mydata \pgfmathsetmacro{\currentY}{\pgfplotsretval}
      \pgfplotstablegetelem{\the\numexpr\idx\relax}{label}\of\mydata \edef\currentLabel{\pgfplotsretval}

      \pgfmathsetmacro{\colorindex}{int(mod(\idx, \numcolors))}
      \ifcase\colorindex\relax
          \xdef\currentcolor{blue!70!black}%
      \or \xdef\currentcolor{red!70!black}%
      \or \xdef\currentcolor{green!60!black}%
      \or \xdef\currentcolor{orange!85!black}%
      \else \xdef\currentcolor{black}%
      \fi


      \pgfmathparse{ifthenelse(\currentX >= 0, "north", "south")} \edef\anchorpos{\pgfmathresult}
      \pgfmathparse{ifthenelse(\currentX >= 0, "12pt", "-15pt")} \edef\xshiftval{\pgfmathresult}

      \draw[color=\currentcolor, dashed, very thick, square tip={\currentcolor}]
           (0,0) -- (\currentX,\currentY);

      \node [anchor=\anchorpos, font=\small, xshift=\xshiftval]
           at (\currentX,\currentY) {\currentLabel};
  } 
\end{tikzpicture}
    \subcaption{Layer $\ell_{8}$} 
  \end{subfigure}
  \hfill
  \begin{subfigure}{0.48\textwidth}
    \centering
\pgfplotstableread[col sep=comma]{atomic_dir_layer_16_llama_1B_W_PCA.csv}{\mydata}
\pgfplotstablegetrowsof{\mydata}
\pgfmathsetmacro{\maxRowIndex}{\pgfplotsretval - 1} 
\begin{tikzpicture}[
    scale=2,                
    square tip/.style args={#1}{ 
        -{Square[length=6.4pt, width=6.4pt, fill=#1]}
    }
]
  \draw[->, gray] (-1.4,0) -- (1.4,0) node[right, black] {$x$};
  \draw[->, gray] (0,-1.4) -- (0,1.4) node[above, black] {$y$};

  \pgfmathsetmacro{\numcolors}{4} 

  \foreach \idx in {0,...,\maxRowIndex} {
      \pgfplotstablegetelem{\the\numexpr\idx\relax}{x}\of\mydata \pgfmathsetmacro{\currentX}{\pgfplotsretval}
      \pgfplotstablegetelem{\the\numexpr\idx\relax}{y}\of\mydata \pgfmathsetmacro{\currentY}{\pgfplotsretval}
      \pgfplotstablegetelem{\the\numexpr\idx\relax}{label}\of\mydata \edef\currentLabel{\pgfplotsretval}

      \pgfmathsetmacro{\colorindex}{int(mod(\idx, \numcolors))}
      \ifcase\colorindex\relax
          \xdef\currentcolor{blue!70!black}%
      \or \xdef\currentcolor{red!70!black}%
      \or \xdef\currentcolor{green!60!black}%
      \or \xdef\currentcolor{orange!85!black}%
      \else \xdef\currentcolor{black}%
      \fi

      \pgfmathparse{ifthenelse(\currentX >= 0, "north", "south")} \edef\anchorpos{\pgfmathresult}
      \pgfmathparse{ifthenelse(\currentX >= 0, "12pt", "-15pt")} \edef\xshiftval{\pgfmathresult}

      \draw[color=\currentcolor, dashed, very thick, square tip={\currentcolor}]
           (0,0) -- (\currentX,\currentY);

      \node [anchor=\anchorpos, font=\small, xshift=\xshiftval]
           at (\currentX,\currentY) {\currentLabel};
  } 
\end{tikzpicture}
    \subcaption{Layer $\ell_{16}$}
  \end{subfigure}

 \caption{\small 2-dimensional \textit{normalized} PCA projection of vectors representing atomic spatial relations \texttt{\{above, below, right, left\}} for the \texttt{Llama-3.2-1B-Instruct} model. (a) Data extracted from layer $\ell_{10}$. (b) Data extracted from layer $\ell_{20}$.}
    \label{fig:all_directions_llama_1B}
\end{figure}

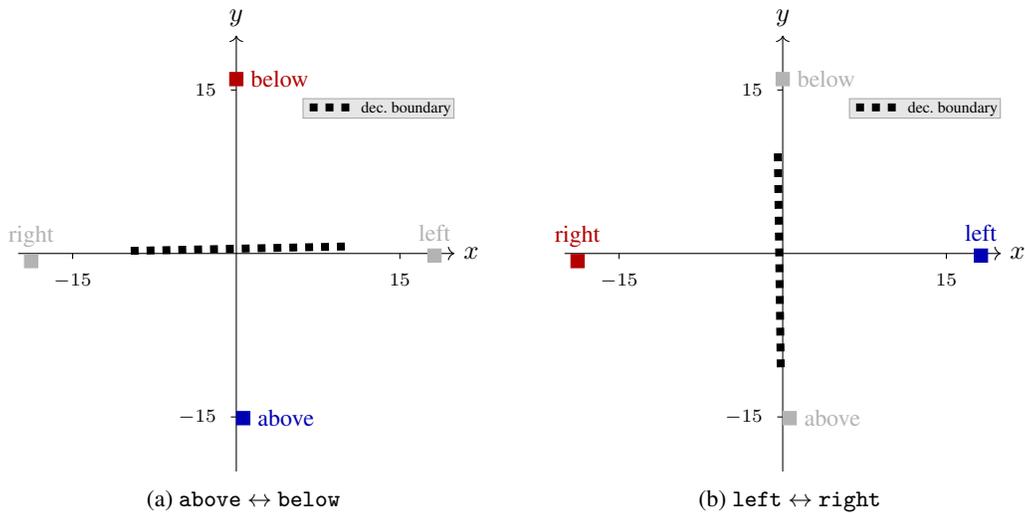
\begin{figure}[h!]
  \centering

  \begin{subfigure}[t]{0.48\textwidth}
    \centering
    \begin{tikzpicture}[scale=.145]
  \node[
    draw=gray!70,           
    fill=gray!20,           
    rectangle,
    inner sep=1pt,
    anchor=north east,
    font=\tiny,
    text=black
  ] at (20.0,14.25) {%
    \raisebox{0.1ex}{%
      \tikz{\draw[dashed,line width=3pt] (0,0) -- (0.6,0);}%
    }\,dec.\ boundary%
  };

  \draw[->] (-20.0,0) -- (20.0,0) node[right] {$x$};
  \draw[->] (0,-20.0) -- (0,20.0) node[above] {$y$};

  \foreach \t in {-15,15}{
    \draw (\t,0) -- (\t,-0.5) node[below=2pt] {\scriptsize $\t$};
    \draw (0,\t) -- (-0.5,\t) node[left=2pt]  {\scriptsize $\t$};
  }

  \pgfplotstableread[col sep=comma,header=true]{boundary_layer_8_llama_1B_above_below.csv}{\data}

  \newcommand{\getxy}[3]{%
    \pgfplotstablegetelem{#1}{x}\of\data
    \pgfmathsetmacro#2{\pgfplotsretval}
    \pgfplotstablegetelem{#1}{y}\of\data
    \pgfmathsetmacro#3{\pgfplotsretval}
  }

  \getxy{4}{\Ax}{\Ay}
  \getxy{5}{\Bx}{\By}
  \draw[line width=3pt,dashed] (\Ax,\Ay) -- (\Bx,\By);

  \def\ms{5pt}
  \foreach \i/\col/\lbl/\anc/\dx/\dy in {
     0/blue!70!black/above/west/ 2pt/ 0pt,
     1/red!70!black/below/west/ 2pt/ 0pt,
     2/gray!60/left/south/ 0pt/ 2pt,
     3/gray!60/right/south/ 0pt/ 2pt
  }{%
    \getxy{\i}{\Px}{\Py}
    \path (\Px,\Py) node[draw=\col,fill=\col,shape=rectangle,
                         minimum size=\ms,inner sep=0pt] {};
    \node[anchor=\anc,font=\small,text=\col,
          xshift=\dx,yshift=\dy]
         at (\Px,\Py) {\lbl};
  }

\end{tikzpicture}
    \subcaption{\texttt{above} $\leftrightarrow$ \texttt{below}}
  \end{subfigure}
  \hfill
  \begin{subfigure}[t]{0.48\textwidth}
    \centering
\begin{tikzpicture}[scale=.145]
  \node[
    draw=gray!70,           
    fill=gray!20,           
    rectangle,
    inner sep=1pt,
    anchor=north east,
    font=\tiny,
    text=black
  ] at (20.0,14.25) {%
    \raisebox{0.1ex}{%
      \tikz{\draw[dashed,line width=3pt] (0,0) -- (0.6,0);}%
    }\,dec.\ boundary%
  };

  \draw[->] (-20.0,0) -- (20.0,0) node[right] {$x$};
  \draw[->] (0,-20.0) -- (0,20.0) node[above] {$y$};

  \foreach \t in {-15,15}{
    \draw (\t,0) -- (\t,-0.5) node[below=2pt] {\scriptsize $\t$};
    \draw (0,\t) -- (-0.5,\t) node[left=2pt]  {\scriptsize $\t$};
  }

  \pgfplotstableread[col sep=comma,header=true]{boundary_layer_8_llama_1B_left_right.csv}{\data}

  \newcommand{\getxy}[3]{%
    \pgfplotstablegetelem{#1}{x}\of\data
    \pgfmathsetmacro#2{\pgfplotsretval}
    \pgfplotstablegetelem{#1}{y}\of\data
    \pgfmathsetmacro#3{\pgfplotsretval}
  }

  \getxy{4}{\Ax}{\Ay}
  \getxy{5}{\Bx}{\By}
  \draw[line width=3pt,dashed] (\Ax,\Ay) -- (\Bx,\By);

  \def\ms{5pt}
  \foreach \i/\col/\lbl/\anc/\dx/\dy in {
     0/gray!60/above/west/ 2pt/ 0pt,
     1/gray!60/below/west/ 2pt/ 0pt,
     2/blue!70!black/left/south/ 0pt/ 2pt,
     3/red!70!black/right/south/ 0pt/ 2pt
  }{%
    \getxy{\i}{\Px}{\Py}
    \path (\Px,\Py) node[draw=\col,fill=\col,shape=rectangle,
                         minimum size=\ms,inner sep=0pt] {};
    \node[anchor=\anc,font=\small,text=\col,
          xshift=\dx,yshift=\dy]
         at (\Px,\Py) {\lbl};
  }

\end{tikzpicture}
    \subcaption{\texttt{left} $\leftrightarrow$ \texttt{right}}
  \end{subfigure}
  \caption{\small Layer~$\ell_{8}$ decision boundaries for inverse relation pairs in \texttt{Llama-3.2-1B-Instruct}, shown in the 2D‐PCA subspace: (a) \texttt{above} $\leftrightarrow$ \texttt{below}, (b) \texttt{left} $\leftrightarrow$ \texttt{right}.}
  \label{fig:llama_1B_decision_boundary_layer8}
\end{figure}

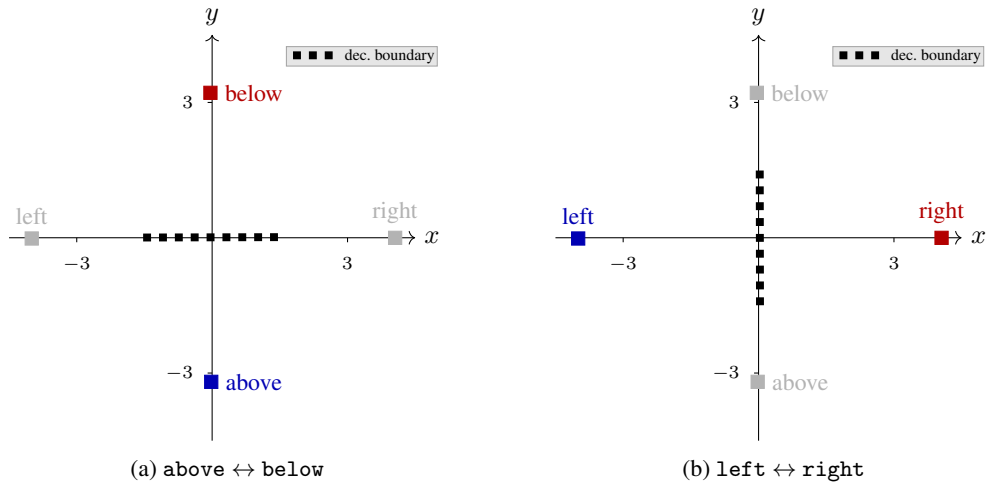
\begin{figure}[h!]
  \centering

  \begin{subfigure}[t]{0.48\textwidth}
    \centering
    \begin{tikzpicture}[scale=.6]
  \node[
    draw=gray!70,           
    fill=gray!20,           
    rectangle,
    inner sep=1pt,
    anchor=north east,
    font=\tiny,
    text=black
  ] at (5.0,4.25) {%
    \raisebox{0.1ex}{%
      \tikz{\draw[dashed,line width=3pt] (0,0) -- (0.6,0);}%
    }\,dec.\ boundary%
  };

  \draw[->] (-4.5,0) -- (4.5,0) node[right] {$x$};
  \draw[->] (0,-4.5) -- (0,4.5) node[above] {$y$};

  \foreach \t in {-3,3}{
    \draw (\t,0) -- (\t,-0.1) node[below=2pt] {\scriptsize $\t$};
    \draw (0,\t) -- (-0.1,\t) node[left=2pt]  {\scriptsize $\t$};
  }

  \pgfplotstableread[col sep=comma,header=true]{boundary_layer_16_llama_1B_above_below.csv}{\data}

  \newcommand{\getxy}[3]{%
    \pgfplotstablegetelem{#1}{x}\of\data
    \pgfmathsetmacro#2{\pgfplotsretval}
    \pgfplotstablegetelem{#1}{y}\of\data
    \pgfmathsetmacro#3{\pgfplotsretval}
  }

  \getxy{4}{\Ax}{\Ay}
  \getxy{5}{\Bx}{\By}
  \draw[line width=3pt,dashed] (\Ax,\Ay) -- (\Bx,\By);

  \def\ms{5pt}
  \foreach \i/\col/\lbl/\anc/\dx/\dy in {
     0/blue!70!black/above/west/ 2pt/ 0pt,
     1/red!70!black/below/west/ 2pt/ 0pt,
     2/gray!60/left/south/ 0pt/ 2pt,
     3/gray!60/right/south/ 0pt/ 2pt
  }{%
    \getxy{\i}{\Px}{\Py}
    \path (\Px,\Py) node[draw=\col,fill=\col,shape=rectangle,
                         minimum size=\ms,inner sep=0pt] {};
    \node[anchor=\anc,font=\small,text=\col,
          xshift=\dx,yshift=\dy]
         at (\Px,\Py) {\lbl};
  }

\end{tikzpicture}
    \subcaption{\texttt{above} $\leftrightarrow$ \texttt{below}}
  \end{subfigure}
  \hfill
  \begin{subfigure}[t]{0.48\textwidth}
    \centering
\begin{tikzpicture}[scale=.6]
  \node[
    draw=gray!70,           
    fill=gray!20,           
    rectangle,
    inner sep=1pt,
    anchor=north east,
    font=\tiny,
    text=black
  ] at (5.0,4.25) {%
    \raisebox{0.1ex}{%
      \tikz{\draw[dashed,line width=3pt] (0,0) -- (0.6,0);}%
    }\,dec.\ boundary%
  };

  \draw[->] (-4.5,0) -- (4.5,0) node[right] {$x$};
  \draw[->] (0,-4.5) -- (0,4.5) node[above] {$y$};

  \foreach \t in {-3,3}{
    \draw (\t,0) -- (\t,-0.1) node[below=2pt] {\scriptsize $\t$};
    \draw (0,\t) -- (-0.1,\t) node[left=2pt]  {\scriptsize $\t$};
  }

  \pgfplotstableread[col sep=comma,header=true]{boundary_layer_16_llama_1B_left_right.csv}{\data}

  \newcommand{\getxy}[3]{%
    \pgfplotstablegetelem{#1}{x}\of\data
    \pgfmathsetmacro#2{\pgfplotsretval}
    \pgfplotstablegetelem{#1}{y}\of\data
    \pgfmathsetmacro#3{\pgfplotsretval}
  }

  \getxy{4}{\Ax}{\Ay}
  \getxy{5}{\Bx}{\By}
  \draw[line width=3pt,dashed] (\Ax,\Ay) -- (\Bx,\By);

  \def\ms{5pt}
  \foreach \i/\col/\lbl/\anc/\dx/\dy in {
     0/gray!60/above/west/ 2pt/ 0pt,
     1/gray!60/below/west/ 2pt/ 0pt,
     2/blue!70!black/left/south/ 0pt/ 2pt,
     3/red!70!black/right/south/ 0pt/ 2pt
  }{%
    \getxy{\i}{\Px}{\Py}
    \path (\Px,\Py) node[draw=\col,fill=\col,shape=rectangle,
                         minimum size=\ms,inner sep=0pt] {};
    \node[anchor=\anc,font=\small,text=\col,
          xshift=\dx,yshift=\dy]
         at (\Px,\Py) {\lbl};
  }

\end{tikzpicture}
    \subcaption{\texttt{left} $\leftrightarrow$ \texttt{right}}
  \end{subfigure}
  \caption{\small Layer~$\ell_{16}$ decision boundaries for inverse relation pairs in \texttt{Llama-3.2-1B-Instruct}, shown in the 2D‐PCA subspace: (a) \texttt{above} $\leftrightarrow$ \texttt{below}, (b) \texttt{left} $\leftrightarrow$ \texttt{right}.}
  \label{fig:llama_1B_decision_boundary_layer16}
\end{figure}

\paragraph{Results for \texttt{Llama-3.2-1B-Instruct}.} For this model, we use the linear probe trained for the layers $\ell_{8}$ and $\ell_{16}$ with $d_\text{model} = 2048$. Table~\ref{table:llama_1B_results} illustrates the cosine similarity and angle between the atomic relations for both the original and PCA space. Figures~\ref{fig:all_directions_llama_1B}, \ref{fig:llama_1B_decision_boundary_layer8}, and \ref{fig:llama_1B_decision_boundary_layer16} show, respectively, all the normalized PCA directions for the atomic relations: \texttt{above}, \texttt{below}, \texttt{left}, and \texttt{right}, and decision boundaries between binary relations (\texttt{above}, \texttt{below}), and  (\texttt{left}, \texttt{right}). The results exhibit inverse and orthogonal relational patterns consistent with those found in the main paper using the \texttt{Llama-3.2-3B} model. 

\begin{table}[h!]
\centering
\caption{Comparison of original vs. PCA space vectors for spatial relations (\texttt{above, below, left, right}) from the \texttt{Llama-3.2-1B-Instruct} model. The PCA projection highlights \textcolor{red}{inverse} (\texttt{above, below} and \texttt{left, right}) and \textcolor{blue}{orthogonal} (\texttt{above, left} and \texttt{below, right}) relationships.}
\label{table:llama_1B_results}
\begin{tabular}{cccccc}
\toprule
\textbf{Layer} & \textbf{Relation} & \multicolumn{2}{c}{\textit{Original Space}} & \multicolumn{2}{c}{\textit{PCA-Projected Space}} \\
\cmidrule(lr){3-4} \cmidrule(lr){5-6}
& & Cosine Sim. & Angle (°)  & Cosine Sim.  & Angle (°)  \\
\midrule
\multirow{3}{*}{8} 
& above $\leftrightarrow$ below & 0.6376 & 50.38 & \textcolor{red}{0.9991} & \textcolor{red}{2.42} \\
& left $\leftrightarrow$ right & 0.7249 & 43.54 & \textcolor{red}{0.9989} & \textcolor{red}{2.73} \\
& above $\leftrightarrow$ left & 0.0699 & 85.99 & \textcolor{blue}{-0.0527} & \textcolor{blue}{93.02} \\
& below $\leftrightarrow$ right & 0.0982 & 84.36 & \textcolor{blue}{0.0371} & \textcolor{blue}{87.87} \\
\midrule
\multirow{3}{*}{16} 
& above $\leftrightarrow$ below & 0.6408 & 50.15 & \textcolor{red}{0.9998} & \textcolor{red}{1.03} \\
& left $\leftrightarrow$ right & 0.8422 & 32.63 & \textcolor{red}{1.0000} & \textcolor{red}{0.27} \\
& above $\leftrightarrow$ left & 0.0606 & 86.53 & \textcolor{blue}{-0.0104} & \textcolor{blue}{90.59} \\
& below $\leftrightarrow$ right & 0.0742 & 85.75 & \textcolor{blue}{0.0123} & \textcolor{blue}{89.30} \\
\bottomrule
\end{tabular}
\end{table}

\begin{figure}[h!]
\centering
    \begin{subfigure}{0.48\textwidth}
    \centering
\pgfplotstableread[col sep=comma]{atomic_dir_layer_16_qwen3_W_PCA.csv}{\mydata}
\pgfplotstablegetrowsof{\mydata}
\pgfmathsetmacro{\maxRowIndex}{\pgfplotsretval - 1} 
\begin{tikzpicture}[
    scale=2,                
    square tip/.style args={#1}{ 
        -{Square[length=6.4pt, width=6.4pt, fill=#1]}
    }
]
  \draw[->, gray] (-1.4,0) -- (1.4,0) node[right, black] {$x$};
  \draw[->, gray] (0,-1.4) -- (0,1.4) node[above, black] {$y$};

  \pgfmathsetmacro{\numcolors}{4} 

  \foreach \idx in {0,...,\maxRowIndex} {
      \pgfplotstablegetelem{\the\numexpr\idx\relax}{x}\of\mydata \pgfmathsetmacro{\currentX}{\pgfplotsretval}
      \pgfplotstablegetelem{\the\numexpr\idx\relax}{y}\of\mydata \pgfmathsetmacro{\currentY}{\pgfplotsretval}
      \pgfplotstablegetelem{\the\numexpr\idx\relax}{label}\of\mydata \edef\currentLabel{\pgfplotsretval}

      \pgfmathsetmacro{\colorindex}{int(mod(\idx, \numcolors))}
      \ifcase\colorindex\relax
          \xdef\currentcolor{blue!70!black}%
      \or \xdef\currentcolor{red!70!black}%
      \or \xdef\currentcolor{green!60!black}%
      \or \xdef\currentcolor{orange!85!black}%
      \else \xdef\currentcolor{black}%
      \fi


      \pgfmathparse{ifthenelse(\currentX >= 0, "north", "south")} \edef\anchorpos{\pgfmathresult}
      \pgfmathparse{ifthenelse(\currentX >= 0, "12pt", "-15pt")} \edef\xshiftval{\pgfmathresult}

      \draw[color=\currentcolor, dashed, very thick, square tip={\currentcolor}]
           (0,0) -- (\currentX,\currentY);

      \node [anchor=\anchorpos, font=\small, xshift=\xshiftval]
           at (\currentX,\currentY) {\currentLabel};
  } 
\end{tikzpicture}
    \subcaption{Layer $\ell_{16}$} 
  \end{subfigure}
  \hfill
  \begin{subfigure}{0.48\textwidth}
    \centering
\pgfplotstableread[col sep=comma]{atomic_dir_layer_20_qwen3_W_PCA.csv}{\mydata}
\pgfplotstablegetrowsof{\mydata}
\pgfmathsetmacro{\maxRowIndex}{\pgfplotsretval - 1} 
\begin{tikzpicture}[
    scale=2,                
    square tip/.style args={#1}{ 
        -{Square[length=6.4pt, width=6.4pt, fill=#1]}
    }
]
  \draw[->, gray] (-1.4,0) -- (1.4,0) node[right, black] {$x$};
  \draw[->, gray] (0,-1.4) -- (0,1.4) node[above, black] {$y$};

  \pgfmathsetmacro{\numcolors}{4} 

  \foreach \idx in {0,...,\maxRowIndex} {
      \pgfplotstablegetelem{\the\numexpr\idx\relax}{x}\of\mydata \pgfmathsetmacro{\currentX}{\pgfplotsretval}
      \pgfplotstablegetelem{\the\numexpr\idx\relax}{y}\of\mydata \pgfmathsetmacro{\currentY}{\pgfplotsretval}
      \pgfplotstablegetelem{\the\numexpr\idx\relax}{label}\of\mydata \edef\currentLabel{\pgfplotsretval}

      \pgfmathsetmacro{\colorindex}{int(mod(\idx, \numcolors))}
      \ifcase\colorindex\relax
          \xdef\currentcolor{blue!70!black}%
      \or \xdef\currentcolor{red!70!black}%
      \or \xdef\currentcolor{green!60!black}%
      \or \xdef\currentcolor{orange!85!black}%
      \else \xdef\currentcolor{black}%
      \fi

      \pgfmathparse{ifthenelse(\currentX >= 0, "north", "south")} \edef\anchorpos{\pgfmathresult}
      \pgfmathparse{ifthenelse(\currentX >= 0, "12pt", "-15pt")} \edef\xshiftval{\pgfmathresult}

      \draw[color=\currentcolor, dashed, very thick, square tip={\currentcolor}]
           (0,0) -- (\currentX,\currentY);

      \node [anchor=\anchorpos, font=\small, xshift=\xshiftval]
           at (\currentX,\currentY) {\currentLabel};
  } 
\end{tikzpicture}
    \subcaption{Layer $\ell_{20}$}
  \end{subfigure}

 \caption{\small 2-dimensional \textit{normalized} PCA projection of vectors representing atomic spatial relations \texttt{\{above, below, right, left\}} for the \texttt{Qwen3-1B} model. (a) Data extracted from layer $\ell_{16}$. (b) Data extracted from layer $\ell_{20}$.}
    \label{fig:all_directions_qwen3}
\end{figure}
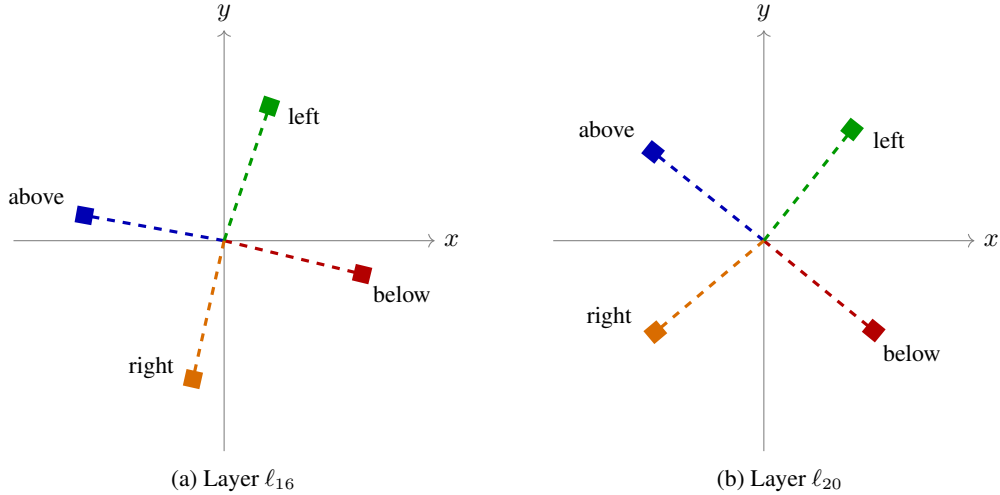

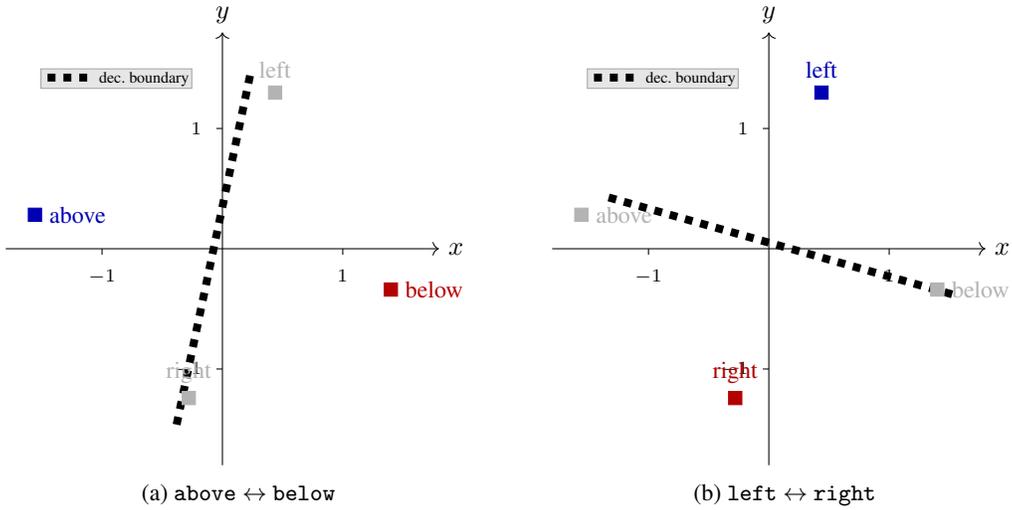
\begin{figure}[h!]
  \centering

  \begin{subfigure}[t]{0.48\textwidth}
    \centering
    \begin{tikzpicture}[scale=1.6]
  \node[
    draw=gray!70,           
    fill=gray!20,           
    rectangle,
    inner sep=1pt,
    anchor=north east,
    font=\tiny,
    text=black
  ] at (-0.25,1.5) {%
    \raisebox{0.1ex}{%
      \tikz{\draw[dashed,line width=3pt] (0,0) -- (0.6,0);}%
    }\,dec.\ boundary%
  };

  \draw[->] (-1.8,0) -- (1.8,0) node[right] {$x$};
  \draw[->] (0,-1.8) -- (0,1.8) node[above] {$y$};

  \foreach \t in {-1,1}{
    \draw (\t,0) -- (\t,-0.05) node[below=2pt] {\scriptsize $\t$};
    \draw (0,\t) -- (-0.05,\t) node[left=2pt]  {\scriptsize $\t$};
  }

  \pgfplotstableread[col sep=comma,header=true]{boundary_layer_16_qwen3_above_below.csv}{\data}

  \newcommand{\getxy}[3]{%
    \pgfplotstablegetelem{#1}{x}\of\data
    \pgfmathsetmacro#2{\pgfplotsretval}
    \pgfplotstablegetelem{#1}{y}\of\data
    \pgfmathsetmacro#3{\pgfplotsretval}
  }

  \getxy{4}{\Ax}{\Ay}
  \getxy{5}{\Bx}{\By}
  \draw[line width=3pt,dashed] (\Ax,\Ay) -- (\Bx,\By);

  \def\ms{5pt}
  \foreach \i/\col/\lbl/\anc/\dx/\dy in {
     0/blue!70!black/above/west/ 2pt/ 0pt,
     1/red!70!black/below/west/ 2pt/ 0pt,
     2/gray!60/left/south/ 0pt/ 2pt,
     3/gray!60/right/south/ 0pt/ 2pt
  }{%
    \getxy{\i}{\Px}{\Py}
    \path (\Px,\Py) node[draw=\col,fill=\col,shape=rectangle,
                         minimum size=\ms,inner sep=0pt] {};
    \node[anchor=\anc,font=\small,text=\col,
          xshift=\dx,yshift=\dy]
         at (\Px,\Py) {\lbl};
  }

\end{tikzpicture}
    \subcaption{\texttt{above} $\leftrightarrow$ \texttt{below}}
  \end{subfigure}
  \hfill
  \begin{subfigure}[t]{0.48\textwidth}
    \centering
\begin{tikzpicture}[scale=1.6]
  \node[
    draw=gray!70,           
    fill=gray!20,           
    rectangle,
    inner sep=1pt,
    anchor=north east,
    font=\tiny,
    text=black
  ] at (-0.25,1.5) {%
    \raisebox{0.1ex}{%
      \tikz{\draw[dashed,line width=3pt] (0,0) -- (0.6,0);}%
    }\,dec.\ boundary%
  };

  \draw[->] (-1.8,0) -- (1.8,0) node[right] {$x$};
  \draw[->] (0,-1.8) -- (0,1.8) node[above] {$y$};

  \foreach \t in {-1,1}{
    \draw (\t,0) -- (\t,-0.05) node[below=2pt] {\scriptsize $\t$};
    \draw (0,\t) -- (-0.05,\t) node[left=2pt]  {\scriptsize $\t$};
  }

  \pgfplotstableread[col sep=comma,header=true]{boundary_layer_16_qwen3_left_right.csv}{\data}

  \newcommand{\getxy}[3]{%
    \pgfplotstablegetelem{#1}{x}\of\data
    \pgfmathsetmacro#2{\pgfplotsretval}
    \pgfplotstablegetelem{#1}{y}\of\data
    \pgfmathsetmacro#3{\pgfplotsretval}
  }

  \getxy{4}{\Ax}{\Ay}
  \getxy{5}{\Bx}{\By}
  \draw[line width=3pt,dashed] (\Ax,\Ay) -- (\Bx,\By);

  \def\ms{5pt}
  \foreach \i/\col/\lbl/\anc/\dx/\dy in {
     0/gray!60/above/west/ 2pt/ 0pt,
     1/gray!60/below/west/ 2pt/ 0pt,
     2/blue!70!black/left/south/ 0pt/ 2pt,
     3/red!70!black/right/south/ 0pt/ 2pt
  }{%
    \getxy{\i}{\Px}{\Py}
    \path (\Px,\Py) node[draw=\col,fill=\col,shape=rectangle,
                         minimum size=\ms,inner sep=0pt] {};
    \node[anchor=\anc,font=\small,text=\col,
          xshift=\dx,yshift=\dy]
         at (\Px,\Py) {\lbl};
  }

\end{tikzpicture}
    \subcaption{\texttt{left} $\leftrightarrow$ \texttt{right}}
  \end{subfigure}
  \caption{\small Layer~$\ell_{16}$ decision boundaries for inverse relation pairs in \texttt{Qwen3-1B}, shown in the 2D‐PCA subspace: (a) \texttt{above} $\leftrightarrow$ \texttt{below}, (b) \texttt{left} $\leftrightarrow$ \texttt{right}.}
  \label{fig:qwen3_decision_boundary_layer16}
\end{figure}

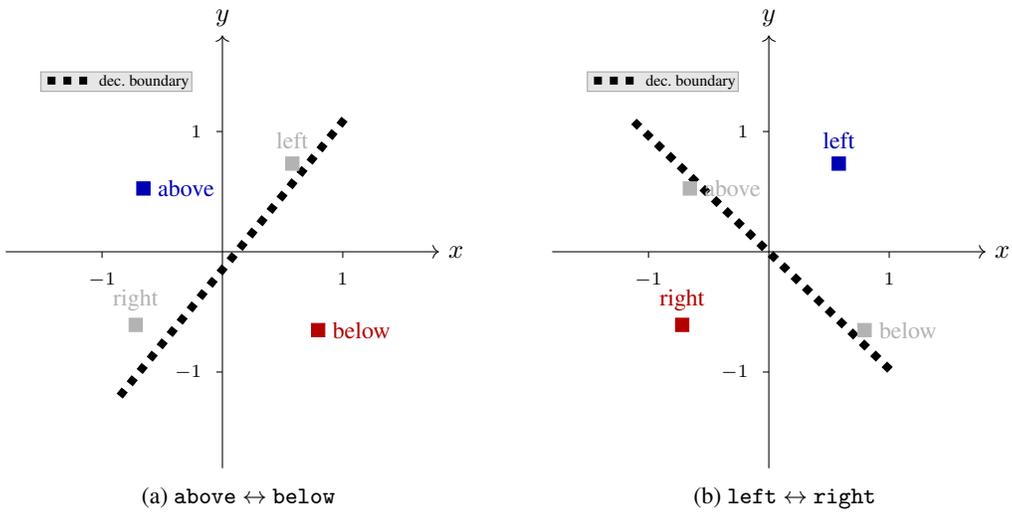
\begin{figure}[h!]
  \centering

  \begin{subfigure}[t]{0.48\textwidth}
    \centering
    \begin{tikzpicture}[scale=1.6]
  \node[
    draw=gray!70,           
    fill=gray!20,           
    rectangle,
    inner sep=1pt,
    anchor=north east,
    font=\tiny,
    text=black
  ] at (-0.25,1.5) {%
    \raisebox{0.1ex}{%
      \tikz{\draw[dashed,line width=3pt] (0,0) -- (0.6,0);}%
    }\,dec.\ boundary%
  };

  \draw[->] (-1.8,0) -- (1.8,0) node[right] {$x$};
  \draw[->] (0,-1.8) -- (0,1.8) node[above] {$y$};

  \foreach \t in {-1,1}{
    \draw (\t,0) -- (\t,-0.05) node[below=2pt] {\scriptsize $\t$};
    \draw (0,\t) -- (-0.05,\t) node[left=2pt]  {\scriptsize $\t$};
  }

  \pgfplotstableread[col sep=comma,header=true]{boundary_layer_20_qwen3_above_below.csv}{\data}

  \newcommand{\getxy}[3]{%
    \pgfplotstablegetelem{#1}{x}\of\data
    \pgfmathsetmacro#2{\pgfplotsretval}
    \pgfplotstablegetelem{#1}{y}\of\data
    \pgfmathsetmacro#3{\pgfplotsretval}
  }

  \getxy{4}{\Ax}{\Ay}
  \getxy{5}{\Bx}{\By}
  \draw[line width=3pt,dashed] (\Ax,\Ay) -- (\Bx,\By);

  \def\ms{5pt}
  \foreach \i/\col/\lbl/\anc/\dx/\dy in {
     0/blue!70!black/above/west/ 2pt/ 0pt,
     1/red!70!black/below/west/ 2pt/ 0pt,
     2/gray!60/left/south/ 0pt/ 2pt,
     3/gray!60/right/south/ 0pt/ 2pt
  }{%
    \getxy{\i}{\Px}{\Py}
    \path (\Px,\Py) node[draw=\col,fill=\col,shape=rectangle,
                         minimum size=\ms,inner sep=0pt] {};
    \node[anchor=\anc,font=\small,text=\col,
          xshift=\dx,yshift=\dy]
         at (\Px,\Py) {\lbl};
  }

\end{tikzpicture}
    \subcaption{\texttt{above} $\leftrightarrow$ \texttt{below}}
  \end{subfigure}
  \hfill
  \begin{subfigure}[t]{0.48\textwidth}
    \centering
\begin{tikzpicture}[scale=1.6]
  \node[
    draw=gray!70,           
    fill=gray!20,           
    rectangle,
    inner sep=1pt,
    anchor=north east,
    font=\tiny,
    text=black
  ] at (-0.25,1.5) {%
    \raisebox{0.1ex}{%
      \tikz{\draw[dashed,line width=3pt] (0,0) -- (0.6,0);}%
    }\,dec.\ boundary%
  };

  \draw[->] (-1.8,0) -- (1.8,0) node[right] {$x$};
  \draw[->] (0,-1.8) -- (0,1.8) node[above] {$y$};

  \foreach \t in {-1,1}{
    \draw (\t,0) -- (\t,-0.05) node[below=2pt] {\scriptsize $\t$};
    \draw (0,\t) -- (-0.05,\t) node[left=2pt]  {\scriptsize $\t$};
  }

  \pgfplotstableread[col sep=comma,header=true]{boundary_layer_20_qwen3_left_right.csv}{\data}

  \newcommand{\getxy}[3]{%
    \pgfplotstablegetelem{#1}{x}\of\data
    \pgfmathsetmacro#2{\pgfplotsretval}
    \pgfplotstablegetelem{#1}{y}\of\data
    \pgfmathsetmacro#3{\pgfplotsretval}
  }

  \getxy{4}{\Ax}{\Ay}
  \getxy{5}{\Bx}{\By}
  \draw[line width=3pt,dashed] (\Ax,\Ay) -- (\Bx,\By);

  \def\ms{5pt}
  \foreach \i/\col/\lbl/\anc/\dx/\dy in {
     0/gray!60/above/west/ 2pt/ 0pt,
     1/gray!60/below/west/ 2pt/ 0pt,
     2/blue!70!black/left/south/ 0pt/ 2pt,
     3/red!70!black/right/south/ 0pt/ 2pt
  }{%
    \getxy{\i}{\Px}{\Py}
    \path (\Px,\Py) node[draw=\col,fill=\col,shape=rectangle,
                         minimum size=\ms,inner sep=0pt] {};
    \node[anchor=\anc,font=\small,text=\col,
          xshift=\dx,yshift=\dy]
         at (\Px,\Py) {\lbl};
  }

\end{tikzpicture}
    \subcaption{\texttt{left} $\leftrightarrow$ \texttt{right}}
  \end{subfigure}
  \caption{Layer~$\ell_{20}$ decision boundaries for inverse relation pairs in \texttt{Qwen3-1B}, shown in the 2D‐PCA subspace: (a) \texttt{above} $\leftrightarrow$ \texttt{below}, (b) \texttt{left} $\leftrightarrow$ \texttt{right}.}
  \label{fig:qwen3_decision_boundary_layer20}
\end{figure}

\textbf{Results for \texttt{Qwen3-1B}.} For this model, we use the linear probe trained for layers $\ell_{16}$ and $\ell_{20}$ with $d_\text{model} = 2048$. Table~\ref{table:qwen3_results} illustrates the cosine similarity and angle between the atomic relations for both the original and PCA space. Figures~\ref{fig:all_directions_qwen3}, \ref{fig:qwen3_decision_boundary_layer16}, and \ref{fig:qwen3_decision_boundary_layer20} show, respectively, all the directions for the atomic relations: \texttt{above}, \texttt{below}, \texttt{left}, and \texttt{right}, and decision boundaries between binary relations (\texttt{above}, \texttt{below}), and  (\texttt{left}, \texttt{right}). As before, the results exhibit inverse and orthogonal relational patterns consistent with those found in the main paper using the \texttt{Llama-3.2-3B} model.

\begin{table}[h!]
\centering
\caption{Comparison of original vs. PCA space vectors for spatial relations (\texttt{above, below, left, right}) from the \texttt{Qwen3-1B} model. The PCA projection highlights \textcolor{red}{inverse} (\texttt{above, below} and \texttt{left, right}) and \textcolor{blue}{orthogonal} (\texttt{above, left} and \texttt{below, right}) relationships.}
\label{table:qwen3_results}
\begin{tabular}{cccccc}
\toprule
\textbf{Layer} & \textbf{Relation} & \multicolumn{2}{c}{\textit{Original Space}} & \multicolumn{2}{c}{\textit{PCA-Projected Space}} \\
\cmidrule(lr){3-4} \cmidrule(lr){5-6}
& & Cosine Sim. & Angle (°)  & Cosine Sim.  & Angle (°)  \\
\midrule
\multirow{3}{*}{16} 
& above $\leftrightarrow$ below & 0.5006 & 59.96 & \textcolor{red}{0.9983} & \textcolor{red}{3.35} \\
& left $\leftrightarrow$ right & 0.3433 & 62.92 & \textcolor{red}{0.9946} & \textcolor{red}{5.94} \\
& above $\leftrightarrow$ left & 0.1201 & 83.10 & \textcolor{blue}{0.1452} & \textcolor{blue}{81.65} \\
& below $\leftrightarrow$ right & 0.1336 & 82.32 & \textcolor{blue}{-0.0165} & \textcolor{blue}{90.95} \\
\midrule
\multirow{3}{*}{20} 
& above $\leftrightarrow$ below & 0.2047 & 78.19 & \textcolor{red}{0.9999} & \textcolor{red}{0.59} \\
& left $\leftrightarrow$ right & 0.1682 & 80.32 & \textcolor{red}{0.9802} & \textcolor{red}{11.43} \\
& above $\leftrightarrow$ left & 0.1024 & 84.12 & \textcolor{blue}{-0.0055} & \textcolor{blue}{90.31} \\
& below $\leftrightarrow$ right & 0.1171 & 83.27 & \textcolor{blue}{0.1827} & \textcolor{blue}{79.47} \\
\bottomrule
\end{tabular}
\end{table}

\end{document}